%% file: energy-planning.tex
\newcommand{\CLASSINPUTtoptextmargin}{19mm}
\newcommand{\CLASSINPUTbottomtextmargin}{19mm}
\newcommand{\CLASSINPUTinnersidemargin}{19mm}
\newcommand{\CLASSINPUToutersidemargin}{19mm}

\documentclass[letterpaper,10pt,conference,twoside]{IEEEtran}
\IEEEoverridecommandlockouts 

\newcommand{\stt}[1]{{\small\tt #1}} 
\newcommand{\powprof}{\stt{powprofiler}}
\newcommand{\figpath}{./figures}

\let\labelindent\relax

\usepackage[inline]{enumitem}
\usepackage{booktabs}
\usepackage{tikz}
\usetikzlibrary{automata,positioning,decorations.pathreplacing,backgrounds}
\usepackage{pifont}

\usepackage{cite}

\usepackage{amsmath}
\makeatletter
\def\maketag@@@#1{\hbox{\m@th\normalfont\normalsize#1}}
\makeatother

\usepackage{mathtools}
\usepackage{amssymb}
\usepackage{arydshln}

\usepackage{amsthm}

\usepackage{algorithm}
\usepackage[noend]{algorithmic}

\usepackage[font=footnotesize]{caption}
\usepackage[font=footnotesize]{subcaption}

\usepackage{url}

\makeatletter
\let\NAT@parse\undefined
\makeatother
\usepackage[pagebackref=true,breaklinks=true,colorlinks,bookmarks=false]{hyperref}
\makeatletter
\newcommand*{\textlabel}[2]{%
  \edef\@currentlabel{#1}
  \phantomsection
  #1\label{#2}
}
\makeatother

\usepackage{textpos}

\DeclarePairedDelimiter\norm{\lVert}{\rVert}%

\theoremstyle{definition}

\newtheorem{defn}{Definition}[section]

\newtheorem*{pb}{Problem}

\DeclareMathOperator*{\argmax}{arg\,max}

\definecolor{c00FFFF}{RGB}{0,255,255}

\begin{document}
\bstctlcite{IEEEexample:BSTcontrol}

\title{\vspace{6mm}\bfseries\LARGE Energy-Aware Planning-Scheduling for Autonomous Aerial Robots}

\author{
  Adam Seewald$^{\text{1}}$, H\'ector Garc\'ia de Marina$^{\text{2}}$, Henrik Skov Midtiby$^{\text{3}}$, and Ulrik Pagh Schultz$^{\text{3}}$
  \thanks{The work was partly funded by EU grant \textnumero 779882 (TeamPlay). The work for H.\hspace*{.4ex}G. is supported by the Ramon y Cajal grant \textnumero RYC2020-030090-I.}
  \thanks{$^{\text{1}}$A.\hspace*{.4ex}S. is with the Department of Mechanical Engineering and Materials Science, Yale University, CT, USA, but the work was performed while affiliated with SDU UAS. Email: {\tt\footnotesize \href{mailto:adam.seewald@yale.edu}{adam.seewald@yale.edu}};}
  \thanks{$^{\text{2}}$H.\hspace*{.4ex}G. is with the Department of Computer Engineering, Automation, and Robotics and with CITIC, University of Granada, Spain;} 
  \thanks{$^{\text{3}}$H.\hspace*{.3ex}S.\hspace*{.3ex}M.,\hspace*{.5ex}U.\hspace*{.3ex}P.\hspace*{.3ex}S. are with SDU UAS, 
  University of Southern Denmark.}
}

\maketitle

\vspace*{-.6ex}
\begin{abstract}
  In this paper, we present an online planning-scheduling approach for battery-powered autonomous aerial robots. The approach consists of simultaneously planning a coverage path and scheduling onboard computational tasks. We further derive a novel variable coverage motion robust to airborne constraints and an empirically motivated energy model. The model includes the energy contribution of the schedule based on an automatic computational energy modeling tool. Our experiments show how an initial flight plan is adjusted online as a function of the available battery, accounting for uncertainty. Our approach 
  remedies possible in-flight failure in case of unexpected battery drops, e.g., due to adverse atmospheric conditions, and increases the overall fault tolerance.
\end{abstract}


%

\vspace*{-.6ex}
\section{Introduction}  %
\begin{figure}[t]
  \centering
  \vspace*{-5ex}
        {\color{black}\scriptsize \input{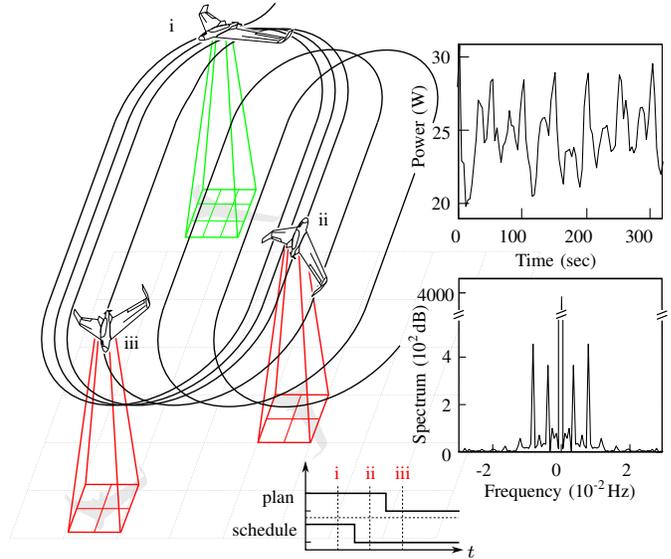}}
  \vspace*{.6ex}
  \caption{
  An initial plan~(in~\ref{sth:i}) {\color{black}is} re-planned online, changing the 
  detection rate or other computational aspects~(in~\ref{sth:ii}) and 
  the number of fly-bys or other motion aspects~(in~\ref{sth:iii}). 
  {\color{black}On the right} 
  {\color{black}are} the energy data of a 
  fixed-wing aerial robot flying a static coverage plan similar to the one illustrated\hspace*{.5ex}here\hspace*{.5ex}{\color{black}and 
  }
  the spectrum analysis, 
  revealing the periodicity 
  exploited in the 
  energy model.
  }
  \label{fig:il-abs}
  \vspace*{-3.8ex}
\end{figure}

\label{sec:intro}       %
Use cases involving aerial robots span broadly. They comprise diverse planning and scheduling strategies and often require high autonomy under strict energy budgets. 
One such use case is coverage path planning (CPP)~\cite{choset2001coverage,galceran2013survey}, which consists of, e.g., an aerial robot visiting every point in a given space~\cite{cabreira2019survey} while running assigned computational tasks. Here, the aerial robot might detect ground patterns and notify other ground-based actors. 
Such use cases arise in 
precision agriculture~\cite{hajjaj2014review} 
where 
{\color{black}information collection prior to a harvesting operation and}
damage prevention during the operation involve aerial robots~\cite{puri2017agriculture, daponte2019review}.
Microcontrollers and heterogeneous computing hardware~\cite{mei2005case} (i.e., with CPUs and GPUs) running power-demanding computational tasks are frequently mounted onto the robots in these and many other scenarios~\cite{william2019aerial
,alexey2021autonomous}.
We refer to onboard computational tasks that can be scheduled with an energy impact as \emph{computations}. We are interested in the energy optimization of motion plans and computations schedules in-flight and refer to it as energy-aware \emph{planning-scheduling}. 
The energy optimization of computations schedules can be achieved by, e.g., varying the quality of service between specific bounds~\cite{ho2019qos} and frequency and voltage of the computing hardware~\cite{mei2005case,brateman2006energy,zhang2007low}. We focus on the former aspect and schedule the onboard computations altering their quality while simultaneously changing the quality of the coverage.
{\color{black} Concretely, we alter how often the aerial robot detects ground patterns along with the distance of the lines that form the coverage.}
Figure~\ref{fig:il-abs} illustrates the intuition: an aerial robot flies a plan with maximal coverage and schedule~(\ref{sth:i}), that is optimized during flight to respect the battery state~(\ref{sth:ii}), and altered due to, e.g., 
battery defects~(\ref{sth:iii}).

There are numerous planning approaches applied to a variety of robots. An instance is an algorithm selecting an energy-optimized trajectory~\cite{mei2004energy} by, e.g., maximizing the operational time~\cite{wahab2015energy}. Many approaches apply to a small number of robots~\cite{kim2005energy} and focus exclusively on planning the trajectory~\cite{kim2008minimum}, despite compelling evidence of the energy influence of onboard {\color{black}computations}~\cite{mei2005case,ondruska2015scheduled,sudhakar2020balancing,brateman2006energy}. In view of the availability of powerful heterogeneous computing hardware~\cite{rizvi2017general}, the use of onboard computations is further expected to increase in the foreseeable future~\cite{
jaramillo2019visual}. In this context, planning-scheduling energy awareness is a recent research direction
~\cite{brateman2006energy,sudhakar2020balancing,lahijanian2018resource,ondruska2015scheduled}. Early studies (2000--2010) varied hardware-dependent aspects, e.g., frequency {\color{black}and} voltage, along with motion aspects, e.g., motor and travel velocities~\cite{mei2005case,brateman2006energy,zhang2007low,sadrpour2013mission} whereas the literature from the past decade derives energy-aware plans-schedules in broader terms. These include simultaneous considerations for planning-scheduling in perception~\cite{ondruska2015scheduled}, localization~\cite{lahijanian2018resource}, navigation~\cite{ho2019qos}, and anytime planning~\cite{sudhakar2020balancing}.
These studies are focused on ground-based robots~\cite{mei2005case,sadrpour2013mission,lahijanian2018resource,ondruska2015scheduled}, yet, aerial robots are particularly affected by energy considerations, as it would be generally required to land to recharge the battery. 
{\color{black}In terms of aerial coverage, past work considers criteria including the completeness of the coverage and resolution~\cite{difranco2015energy}, and deals with aspects such as the quality of the cover~\cite{difranco2016coverage}, but neglects the energy expenditure of computations and favors rotary-wing aerial robots rather than aerial robots broadly.} 
Such a state of practice has prompted us to propose the planning-scheduling approach for autonomous aerial robots, combining the past body of knowledge but addressing aerial robots' peculiarities such as the atmospheric, battery, and turning radius constraints. Numerical simulations and experimental data of 
static and dynamic plans and schedules show improved power savings and fault tolerance with the 
robot 
remedying in-flight failures. 
 
Our focus is on fixed wings, i.e., airborne robots where wings provide lift, propellers provide forward thrust, and control surfaces perform maneuvering. Here, motion and computations energies are within an order of magnitude from each other~\cite{seewald2020mechanical,zamanakos2020energy}. 
{\color{black}T}here are other classes where planning-scheduling energy awareness leads to irrelevant savings, i.e., when the motion energy contribution far outreaches the computations or vice-versa. The {\color{black}motion outreaching computation energy} frequently happens with rotary-wing aerial robots (e.g., quadrotors or quadcopters, hexacopters, etc.){\color{black}, the opposite occurs with} lighter-than-air aerial robots (e.g., blimps). It is 
common 
{\color{black}in} 
planning-scheduling literature, focusing on 
efficient ground-based robots such as Pioneer~3DX~\cite{ho2019qos,mei2005case}, ARC~Q14~\cite{ondruska2015scheduled,lahijanian2018resource}, and Pack-Bot~UGV~\cite{sadrpour2013mission}.

To guarantee energy awareness, our approach uses optimal control {\color{black} and heuristics} where both the paths and schedules variations are trajectories, varying between given bounds (i.e., physical constraints of the robot and computing hardware, quality of service, desired quality of the coverage, etc.). Past planning-scheduling studies also employ optimization techniques~\cite{brateman2006energy,zhang2007low,ondruska2015scheduled,lahijanian2018resource}; some use a greedy approach~\cite{mei2005case,sudhakar2020balancing,sadrpour2013mission}; whereas others use reinforcement learning-based approaches~\cite{ho2019qos,ho2018towards}. {\color{black}Hybrid approaches~\cite{ondruska2015scheduled} are also available, where the techniques are mixed.} Both the path{\color{black}s} and schedules variations trajectories are derived for future time instants employing computations and overall energies and battery models. The energy model for the computations uses regressional analysis from our earlier study on heterogeneous computing hardware
~\cite{seewald2019coarse,seewald2019component}, whereas the battery uses an equivalent circuit model (ECM) from the literature~\cite{
hinz2019comparison,mousavi2014various}. The overall model wraps these two aspects in a cohesive model that uses dynamics modeling to predict the energy behavior of future plans and schedules. In Fig.~\ref{fig:il-abs}, collected energy data ({\color{black}top-right}) and spectrum analysis ({\color{black}below}) of a fixed wing 
flying CPP motivate the overall energy model: the evolution is periodic--CPP often involves repetitive motions to cover the space~\cite{choset2001coverage,galceran2013survey}--an observation exploited in Section~\ref{sec:energy-model}.

The 
{\color{black}remainder is} then organized as follows. Sec.~\ref{sec:prob} provides basic constructs
{ \color{black}and} 
Sec.~\ref{sec:algo} describes 
the methodology of planning-scheduling. Sec.~\ref{sec:experimental} presents the results 
and Sec.~\ref{sec:conclusion} concludes and provides future perspectives. 

\section{Problem Formulation}  %
\label{sec:prob}               %
                               %
{\color{black}W}e assume 
{\color{black}the }robot contains a \emph{plan} composed of \emph{stages}. At each, 
{\color{black}it }travels a path and runs a schedule on the computing hardware. Both are 
altered in Sec.~\ref{sec:algo} within given boundaries with \emph{path}- and \emph{computation}-specific \emph{parameters}.

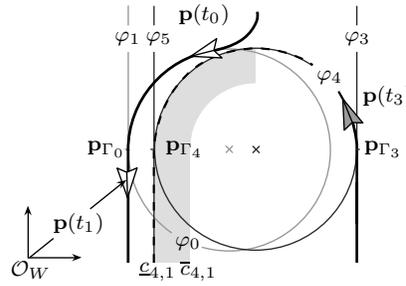
\begin{figure}[t]
  \footnotesize
  \begin{minipage}[t]{0.65\columnwidth}
    \centering
    \input{\figpath/new/traj1.tikz}
  \end{minipage}
  \begin{minipage}[t]{0.012\columnwidth}
  \end{minipage}\hfill
  \begin{minipage}[t]{0.328\columnwidth}
    \vspace*{-28.2ex}
    \caption{Definitions~\ref{def:stage}--\hyperref[def:plan]{4} on a slice of the plan $\Gamma$. $\mathbf{p}_{\Gamma_i}$ are triggering points in which proximity happens {\color{black}the} change of stages $\Gamma_i$. Each contains a path function $\varphi_i$ and parameters to alter the path and schedule $c_{i,1},\dots$.}
    \label{fig:traj1}
  \end{minipage}
  \vspace*{-4ex}
\end{figure}

\subsection{Preliminaries}
\label{sec:prelim}


\begin{defn}[Stage]\label{def:stage}
  Given a generic point $\mathbf{p}\in\mathbb{R}^2$ w.r.t. a reference frame $\mathcal{O}_W$ of the aerial robot flying at a given altitude $h\in\mathbb{R}_{>0}$, the $i$th \emph{stage} $\Gamma_i$ 
  is
  \begin{equation*}\begin{split}
    \Gamma_i:=\{{\color{black}\varphi_i(\mathbf{p}
    ,c_i^\rho)},c_i^\sigma\mid
    \,&\forall j\in\,[\rho]_{>0},\,c_{i,j}\,\,\,\,\,\,\,\in\mathcal{C}_{i,j},\,\\
      &\forall k\in[\sigma]_{>0},\,c_{i,\rho+k}\in\mathcal{S}_{i,k}\,\},
  \end{split}\end{equation*}
  where $c_i^\rho${\color{black}$:=\{c_{i,1},c_{i,2},\dots,c_{i,\rho}\}$} and $c_i^\sigma${\color{black}$:=\{c_{i,\rho+1},c_{i,\rho+2},$ $\dots,c_{i,\rho+\sigma}\}$} are $\rho$ \emph{path} and $\sigma$ \emph{computation parameters}{\color{black}, e.g., $c_i^\rho:=\{c_{i,1}\}$ is a value that changes the distance of the coverage lines and $c^\sigma_i:=\{c_{i,2}\}$ the detection rate with $\rho$ and $\sigma$ being one (see Sec.~\ref{sec:experimental})}. $\mathcal{C}_{i,j}:=[\underline{c}_{i,j},\overline{c}_{i,j}]\subseteq\mathbb{R}$ is the $j$th path parameter 
  constraint set, 
  $\mathcal{S}_{i,k}:=[\underline{c}_{i,\rho+k},\overline{c}_{i,\rho+k}]\subseteq\mathbb{Z}_{\geq 0}$ 
  the $k$th computation parameter constraint set. {\color{black}Indices $j,k$ serve~to~differentiate path and computation parameters constraints and indicate that each parameter can have a different constraint set.} 
\end{defn}

For a set $\mathbb{X}$, $\mathbb{X}_{\geq 0}$ indicates its members are positive, $\mathbb{X}_{> 0}$ strictly positive, and $|\mathbb{X}|$ its cardinality. $\mathbb{Z},\mathbb{R}$ are 
integers and reals. 
Bold letters indicate vectors. 
The notation $[x]$ denotes positive naturals up to $x$, i.e., $\{0,1,\dots,x\}$, {\color{black}$[x]_{>0}$ strictly positive naturals, i.e., $\{1,2,\dots,x\}$,} $x'$ the transpose of $x$, and $[\underline{x},\overline{x}]$ the upper/lower bounds of 
$x$, i.e.,
  $\underline{x}\leq x\leq\overline{x}$.

The function $\varphi_i$ is a \emph{path function}{\color{black}--a }
stage-dependent mathematical function 
the 
robot tracks as it travels 
the coverage. 

\begin{defn}[Path functions]
  \label{def:paths}
  $\varphi_i:\mathbb{R}^2\times\mathbb{R}^\rho\rightarrow\mathbb{R},\,\forall i\in\{1,2,\dots\}
  $ are \emph{path functions}, forming the path. They are a function of {\color{black}$\mathbf{p}
  $} and path parameters $c_i^\rho
  $ and are continuous.
\end{defn}

The change of stages happens in the proximity of given points termed \emph{triggering points}, whereas the plan is complete at the occurrence of the \emph{final point}.

\begin{defn}[Triggering and final points]
  \label{def:trigs}
  The \emph{triggering point} $\mathbf{p}_{\Gamma_{i}}$ describes the transition between stages. \emph{Final point} is the last triggering point $\mathbf{p}_{\Gamma_{l}}$ relative to the last stage $\Gamma_l$.
\end{defn}

The plan merges the concepts from Definitions~\ref{def:stage}--\hyperref[def:trigs]{3}. 

\begin{defn}[Plan]\label{def:plan}
  The \emph{plan} is a finite state machine (FSM) $\Gamma$, where the state-transition function $s:\bigcup_i{\Gamma_i}\times\mathbb{R}^2\rightarrow\bigcup_i{\Gamma_i}$ maps a stage and a point to the next stage
  \begin{equation*}{\color{black}s(\Gamma_i,\mathbf{p}
    )}:=\begin{cases}
    \Gamma_{i+j} & {\color{black}\text{if }\norm{\mathbf{p}
    -\mathbf{p}_{\Gamma_i}}<\varepsilon_i,\,\exists j\in\mathbb{Z},}\\
    \Gamma_i & \text{otherwise}.
  \end{cases}\end{equation*}
\end{defn}

The stage-dependent value $\varepsilon_i\in\mathbb{R}_{\geq 0}$ in Definition~\ref{def:plan} expresses the radius of a non-existent circle over $\mathbf{p}_{\Gamma_i}$.

Fig.~\ref{fig:traj1} illustrates the concepts in Definitions~\ref{def:stage}--\hyperref[def:plan]{4}. $\varphi_0,\dots,\varphi_5$ are path functions. $\varphi_0$ and $\varphi_4$ are circles, while $\varphi_1$, $\varphi_3$, and $\varphi_5$ are lines. They are relative to different stages $\Gamma_1,\dots$ but $\Gamma_0$ (the starting stage) and are changed in the proximity of $\mathbf{p}_{\Gamma_0},\dots$. 
It is possible to alter the paths $\varphi_1,\dots,\varphi_4$ with the parameters $c_{1,1},\dots,c_{4,1}$--
the gray area.

A convenient way of defining $\Gamma$ is specifying a set of stages, a shift, and a final point. The set is termed \emph{primitive stages} and iterated with the shift up to 
the final point.

\begin{defn}[Primitive stages]
  \label{def:primitive}
  Given the number of \emph{primitive stages} $n\in\mathbb{Z}_{>0}$, a \emph{shift} $\mathbf{d}\in\mathbb{R}^2$, and a final point $\mathbf{p}_{\Gamma_l}$, the stages $\Gamma_1,\Gamma_2,\dots,\Gamma_n$ 
  are \emph{primitive} if they form the remainder of the plan with $\mathbf{d}$ up to $\mathbf{p}_{\Gamma_l}$. 
\end{defn}
\noindent In this case, the path functions have a constant distance $e_j$ per each value in $[n]_{>0}$, i.e., 
\begin{equation}\small\label{eq:primitive}
  \varphi_{(i-1)n+j}(\mathbf{p}+(i-1)\mathbf{d},c_1^\rho)-\varphi_{in+j}(\mathbf{p}+i\mathbf{d},c_1^\rho)=e_j,
\end{equation}
holds $\forall i\in[l/n-1]_{>0},j\in[n]_{>0}$ assuming the total number of stages is known and is $l\in\mathbb{Z}_{>0}$. $e_j\in\mathbb{R}$ given a shift $\mathbf{d}$, initial point $\mathbf{p}$, and initial value of path parameters $c_1^\rho$.

\vspace*{-.7ex}
\subsection{Energy-aware planning-scheduling problem}
\label{sec:pbfor}

The problem 
is {\color{black}split into }
{\color{black}the derivation }
{\color{black} of} a {\color{black}coverage} plan 
{\color{black}and its} {\color{black}energy-aware} 
re-plan{\color{black}ing}~and~-schedul{\color{black}ing~}
in-flight.~The 
{\color{black} 
re-planning-scheduling increases a unitless performance metric--
the weighted average of parameters divided by the remaining battery state of charge (SoC) at the end of the flight, both in percent (e.g., $\underline{c}_{i,j}$, $\overline{c}_{i,j}$ correspond to 0 and 100). The objective is~
high~average~parameters configuration and battery usage with successful 
coverage.
}

\begin{pb}[Coverage and re-planning-scheduling problem]
  \label{pb:cov-pb}
  Consider a finite set of vertices of a polygon $v:=\{v_1,v_2,\dots\}$ where each 
  is a point w.r.t. $\mathcal{O}_W$. 
  Let $\underline{r}\in\mathbb{R}_{\geq 0}$, the vehicle's turning radius, and $\mathbf{p}(t_0)$, the starting point at the time instant $t_0$, be given. 
  The \emph{coverage problem} is the problem of finding a plan $\Gamma$ to cover the polygon, whereas the \emph{re-planning-scheduling problem} is finding the {\color{black}energy-aware} trajectory of parameters $c_i$ in time{\color{black}.}
\end{pb}    



Here, $c_i$ denotes a row vector with both the path and computation parameters in sequence, i.e., $c_i:=[\begin{matrix}\,c_i^\rho & c_i^\sigma\,\end{matrix}]'$. 

\section{Energy Models}  %
\label{sec:energy-model} %
The solution to the problem requires energy models, predicting the impact of changes to path and computation parameters on the battery. 
Sec.~\ref{sec:mod-mot}--\hyperref[sec:mod-bat]{C} {\color{black}thus} provide models for the overall and computations energies {\color{black}and }
battery evolution.

\subsection{Overall energy model
}
\label{sec:mod-mot}

The collected energy data and corresponding spectrum analysis in Fig.~\ref{fig:il-abs} show the energy of a static coverage plan. It is relative to one flight of a series of flights for CPP 
in a precision agriculture use case~\cite{seewald2020mechanical}. 
Assuming the primitive paths have approximately the same length and the aerial robot has a fixed ground speed, the data exhibits periodic behavior with a constant set of frequencies, independent of the shift. The hypothesis is further backed by the power spectrum analysis, indicating that to model the energy, three frequencies are adequate.




An intuitive way of modeling the energy data is a Fourier series of a given order $r\in\mathbb{Z}_{\geq 0}$ and period $T\in\mathbb{R}_{>0}$
\begin{equation}\label{eq:fourier}
  h(t)=a_0/T+(2/T)\sum_{j=1}^{r}{\left(a_j\cos{\omega jt}+b_j\sin{\omega jt}\right)},
\end{equation}
where $h:\mathbb{R}_{\geq 0}\rightarrow\mathbb{R}$ maps time to the instantaneous energy
, $\omega:=2\pi/T$ is the angular frequency, and $a,b\in\mathbb{R}$ 
coefficients.

Equation~(\ref{eq:fourier}) does not account for the variation of parameters, where, e.g., two schedules result in different instantaneous energies.
For this latter purpose, we use the dynamics 
\begin{subequations}\label{eq:state-perf}\vspace*{-3ex}
  \begin{align}
  \dot{\mathbf{q}}(t)&=A\mathbf{q}(t)+B\mathbf{u}(t),\label{eq:state-perf-q}\\
  y(t)&=C\mathbf{q}(t),\label{eq:state-perf-y}
\end{align}
\end{subequations}
where $y(t)\in\mathbb{R}$ is the instantaneous energy consumption. The state $\mathbf{q}\in\mathbb{R}^m$ with $m:=2r+1$ contains energy coefficients
\begin{equation}
  \mathbf{q}(t)=\begin{bmatrix}
    \alpha_0(t) & \alpha_1(t) & \beta_1(t) & \cdots & \alpha_r(t) & \beta_r(t)
  \end{bmatrix}'.
\end{equation}

The state transition matrix
\begin{equation}\label{eq:mat_A}{\small
  A=\begin{bmatrix}
    0            & 0^{1\times 2}& \dots & 0^{1\times 2} \\
    0^{2\times 1}& A_1          & \dots & 0^{2\times 2} \\
    \vdots       & \vdots       & \ddots& \vdots        \\
    0^{2\times 1}& 0^{2\times 2}& \dots & A_r 
  \end{bmatrix},\,\,A_j:=\begin{bmatrix}0 & \omega j \\ -\omega j & 0\end{bmatrix}},
\end{equation}
where $A\in\mathbb{R}^{m\times m}$ contains $r$ sub-matrices $A_j$ and $0^{i\times j}$ is a zero matrix of $i$ rows and $j$ columns. In matrix $A$, the top left entry is zero, the diagonal entries are $A_1,\dots,A_r$, the remaining entries are zeros.

The output matrix\vspace*{-3ex}
\begin{equation}\label{eq:mat_C}
  C=(1/T)\Big[1 \,\,\, \overbrace{\begin{matrix}1 & 0 &\cdots & 1 & 0\end{matrix}}^{2r}\Big],
\end{equation}
where $C\in\mathbb{R}^m$ (the first value in the first column is one, the pattern one--zero is then repeated $2r$ times).



To define the nominal control and the output matrix, we exploit the effect of variation of path and computation parameters on the energy. 
  Given $c_i(t)$ parameters at two following time instants $t\in\{t_j,t_{j+1}\}\subset\mathbb{R}_{\geq 0}$ s.t. $t_j<t_{j+1}$ for an arbitrary stage $\Gamma_i$, a change in parameters $c_i(t_j)\neq c_i(t_{j+1})$ results in different overall and instantaneous energies for path and computation parameters respectively.


%
The nominal control and input matrix in Eq.~(\ref{eq:state-perf}) simply includes the change in energy for all time instants, i.e.,
\begin{equation}\label{eq:mat_B}{\small
  \mathbf{u}(t_{j+1})\hspace*{-.5ex}:=\hspace*{-.5ex}\hat{\mathbf{u}}(t_{j+1})\hspace*{-.5ex}-\hspace*{-.5ex}\hat{\mathbf{u}}(t_j),\,\,\,B=\begin{bmatrix}
      0^{1\times\rho} & 1      & \cdots & 1      \\
      0^{1\times\rho} & 0      & \cdots & 0      \\ 
      \vdots          & \vdots & \ddots & \vdots \\
      0^{1\times\rho} & 0      & \cdots & 0   
  \end{bmatrix}},
\end{equation}
shifts the base frequency $\alpha_0$ assuming the energy of the computations does not alter the other frequencies. $B\in\mathbb{R}^{m\times n}$ with $n:=\rho+\sigma$ contains zeros but in the first row where the first $\rho$ columns are zeros and the remaining $\sigma$ are ones. Different combinations of $\mathbf{u}$ with matrix $B$ in Eq.~(\ref{eq:mat_B}) are possible 
{\color{black}(see }Sec.~\ref{sec:conclusion}{\color{black})}.
The dynamics in Eq.~(\ref{eq:state-perf}--\ref{eq:mat_B}) additionally allow us to use state estimation techniques, such as the Kalman filter in Sec.~\ref{sec:repla-algo}, to refine the states $\mathbf{q}$ and model the energy of the aerial robot flying under diverse 
conditions.

{\color{black}Matrices $A$ and $C$ are constructed such that t}he 
models in Eq.~(\ref{eq:fourier}--\ref{eq:state-perf}) are equal
when $\mathbf{u}$ is a zero vector 
and an initial guess $\mathbf{q}(t_0)=\mathbf{q}_0$ at {\color{black}the} initial time instant $t_0$
\begin{equation}
  \mathbf{q}_0=\begin{bmatrix}a_0 & a_1/2 & b_1/2 & \cdots & a_r/2 & b_r/2\end{bmatrix}',
\end{equation}
i.e., $h,y$ are 
harmonic signals with the same frequencies{\color{black}. For further details see the first author's Ph.D. thesis~\cite{seewaldphdthesis}}.

$\hat{\mathbf{u}}$ in Eq.~(\ref{eq:mat_B}) is then a scale transformation
\begin{equation}
  \hat{\mathbf{u}}(t):=\mathrm{diag}(\nu_i)c_i(t)+\tau_i,
\end{equation}
where $\mathrm{diag}(x)$ is a diagonal matrix with items of a set $x$ on the diagonal and zeros elsewhere. $\nu_i:=\begin{bmatrix}\nu_{i,1}&\cdots&\nu_{i,n}\end{bmatrix}'$ and $\tau_i:=\begin{bmatrix}\tau_{i,1}&\cdots&\tau_{i,n}\end{bmatrix}'$ are scaling factors{\color{black},} 
transform{\color{black}ing} parameters 
(see Definition~\ref{def:stage}) to time and power domains.

{\color{black}W}e assume that the coverage time evolves linearly{ \color{black}and that the }path parameters{ \color{black}contribute to it equally.} $c_i^\rho$ can be {\color{black}then} transformed into a time measure with scaling factors\begin{subequations}
  \label{eq:scale-traj}\begin{align}
  \nu_{i,j}&=\left((\overline{t}-\underline{t})/(\overline{c}_{i,j}-\underline{c}_{i,j})\right)/\rho,\\
  \tau_{i,j}&=\left(\underline{c}_{i,j}(\underline{t}-\overline{t})/(\overline{c}_{i,j}-\underline{c}_{i,j})+\underline{t}\right)/\rho,
\end{align}\end{subequations} 
$\forall j\in[\rho]_{>0}$ where $\overline{t},\underline{t}$ are time measures needed to complete the coverage with configurations $\underline{c}_i^\rho,\overline{c}_i^\rho$ ($\underline{\Gamma},\overline{\Gamma}$).

Similarly to Eq.~(\ref{eq:scale-traj}), computation parameters $c_i^\sigma$ can be transformed into an instantaneous energy measure with 
\begin{subequations}
  \label{eq:scale-comp}\begin{align}
  \nu_{i,j}&=(g(\overline{c}_{i,j})-g(\underline{c}_{i,j}))/(\overline{c}_{i,j}-\underline{c}_{i,j}),\\
  \tau_{i,j}&=\underline{c}_{i,j}(g(\underline{c}_{i,j})-g(\overline{c}_{i,j}))/(\overline{c}_{i,j}-\underline{c}_{i,j})+g(\underline{c}_{i,j}),
\end{align}\end{subequations}
$\forall j\in[\rho+1,n]$. The function $g$ is detailed in Sec.~\ref{sec:mod-com} and quantifies the power of the computing hardware.

\vspace*{.4ex}
\subsection{Energy model for the computations}
\label{sec:mod-com}

Models for heterogeneous computing hardware in the literature often rely on analytical expressions~\cite{marowka2017energy,
yang2017designing} or different techniques, such as regressional analysis~\cite{bailey2014adaptive,ma2012holistic,seewald2019coarse}, aiding the selection of hardware- or software-specific parameters. This section presents an energy model based on our early studies~\cite{seewald2019component,seewald2019coarse}, which relies on regressional analysis to quantify the computations energy of any configuration of computations $c_i^\sigma$ within the bounds (see Definition~\ref{def:stage}).

The model compromises a 
modeling and profiling tool~\cite{seewald2019coarse} named \powprof{} distributed 
under the open-source MIT license. It is segmented into two layers. In the \emph{measurement layer}, the tool measures a discrete set of computation parameters and infers the energy of the remaining in the \emph{predictive layer} via a piecewise linear regression.

We assume there is at least one measuring device, i.e., shunt or internal power resistor, multimeter, or amperemeter, quantifying the power drain of a 
component, e.g., CPU, GPU, memory, etc., or of the entire computing hardware.

\begin{defn}[Measurement layer]\label{def:meas}
  Given a measuring device, computation parameters, and initial and final time instants, the \emph{measurement layer} is the function $\gamma:\mathbb{Z}_{>0}\times\mathbb{Z}^\sigma\times\mathcal{T}\rightarrow\mathbb{R}$ that returns an energy measure.
\end{defn}

The notation $\mathcal{T}$ encloses all the time intervals from initial $t_0$ to final $t_f$, i.e., $\mathcal{T}:=[t_0,t_f]$.

\begin{defn}[Predictive layer]\label{def:pred}
  Given a measuring device and computation parameters, the \emph{predictive layer} is the function $g:\mathbb{Z}_{>0}\times\mathbb{Z}^\sigma\rightarrow\mathbb{R}$ that returns an energy measure.
\end{defn}

The energy measures in Definitions~\ref{def:meas}--\hyperref[def:pred]{2} can be either average expressed in watts or overall expressed in joules. Additionally, the \powprof{} tool supports the battery SoC detailed in Sec.~\ref{sec:mod-bat}. The function $g$ in Definition~\ref{def:pred} is contained in the 
factors in Eq.~(\ref{eq:scale-comp}), assuming the computations energy behaves linearly between $\underline{c}_i^\sigma$ and $\overline{c}_i^\sigma$, otherwise
\begin{equation}
  \label{eq:piece-wise-reg}\begin{split}
  g(c_i^\sigma)=(&\gamma(\lceil c_i^\sigma\rceil,\mathcal{T}_1)-\gamma(\lfloor c_i^\sigma\rfloor,\mathcal{T}_2))\\(&c_i^\sigma-\lfloor c_i^\sigma\rfloor)/(\lceil c_i^\sigma\rceil-\lfloor c_i^\sigma\rfloor)+\gamma(\lfloor c_i^\sigma\rfloor,\mathcal{T}_2),
\end{split}\end{equation}
where notation $\lceil c_i^\sigma\rceil,\lfloor c_i^\sigma\rfloor$ indicates two adjacent measurement layers, and $\mathcal{T}_1,\mathcal{T}_2$ 
the corresponding two time intervals. {\color{black}The m}easuring device in $\gamma$ and $g$ is {\color{black}not 
stated in Eq.~(\ref{eq:piece-wise-reg})}.

\subsection{Battery model}
\label{sec:mod-bat}

The battery model predicts the battery SoC as a function of a given load at future time 
instants. There are multiple models in the literature~\cite{rao2003battery} with varying complexity {\color{black}and} accuracy 
ranging from accurate but costly physical models~\cite{
marcicki2013design}, to abstract models~\cite{
hinz2019comparison,mousavi2014various} 
{\color{black}with} compelling trade-offs in terms of the latter two. 
We model a Li-ion battery 
in-flight with an abstract ``Rint'' ECM in the literature~\cite{
hinz2019comparison,mousavi2014various}.

The battery SoC changes according to~\cite{hasan2018exogenous
}, i.e.,
\begin{equation}\label{eq:batdyn}
  \dot{b}(y(t))=-k_bI(y(t))/Q_c,
\end{equation}
where $I(y(t))\hspace*{-.5ex}\in\hspace*{-.5ex}\mathbb{R}$ is the internal current measured~in~am- peres, $y(t)\hspace*{-.5ex}\in\hspace*{-.5ex}\mathbb{R}_{\geq 0}$ the power drain, and $Q_c\hspace*{-.5ex}\in\hspace*{-.5ex}\mathbb{R}$ the battery constant nominal capacity measured in amperes per hour. $k_b$ is a battery coefficient added to~\cite{hasan2018exogenous
} and derived experimentally. The ``Rint'' circuit models the battery as a perfect voltage source connected with a resistor $R_r\in\mathbb{R}$ measured in ohm, representing the battery resistance. The voltage on the extremes of ECM respects $V_e=V-R_rI$, where $V,V_e\in\mathbb{R}$ are the internal and external battery voltages measured in volts. The former can be retrieved from the battery data sheet~\cite{hinz2019comparison} and depends on the SoC~\cite{hasan2018exogenous}.

If the voltage 
is stable, Kirchhoff's circuit laws lead to $V_sI_l=V_eI$, where $I_l$ is the current required by the load 
in amperes. Combining $V_e,V_sI_l$ results in the 
expression $R_rI^2-$ $VI+V_sI_l=0$. Solving the expression utilizing the negative solution (when $I_l$ is zero, $I$ should also be zero) 
results in \vspace*{-1ex}
\begin{equation}\label{eq:batdyn2}
I(y(t))=(V-\sqrt{V^2-4R_ry(t)})/(2R_r).
\end{equation} 
%


Eq.~(\ref{eq:state-perf}) states that the output $y$ evolves in $\mathbb{R}$, 
yet, aerial robots usually use a battery.
We thus use instead
\begin{equation}\label{eq:output-const}
  {
  \mathcal{Y}(t):=\{y\mid y\in[0,b\,Q_cV]\subseteq{\mathbb{R}_{\geq 0}}\},} 
\end{equation}
where $b\,Q_cV$, the maximum instantaneous energy 
measured in watts, is derived from Eq{\color{black}}.~(\ref{eq:batdyn}--\ref{eq:batdyn2}), i.e., the computation parameters in Algorithm~\ref{alg2} and Eq.~(\ref{eq:ocp-output-mpc}) later in Sec.~\ref{sec:repla-algo} will have an energy constraint.
\section{Planning-Scheduling}  %
\label{sec:algo}               %
\begin{figure}[t]
  \footnotesize
  \begin{minipage}[c]{0.23\columnwidth}
    \vspace*{-6.6ex}
    \caption{Change of the path parameter $c_{i,1}$, the radius of the circle (i.e., the alteration of the plan in Fig.~\ref{fig:il-abs}).}
    \label{fig:tee1}
  \end{minipage}\hfill
  \begin{minipage}[c]{0.05\columnwidth}
  \end{minipage}\hfill
  \begin{minipage}[c]{0.71\columnwidth}
    \centering
    \input{\figpath/new/traj2.tikz}
  \end{minipage}
  \vspace*{-4.2ex}
\end{figure}
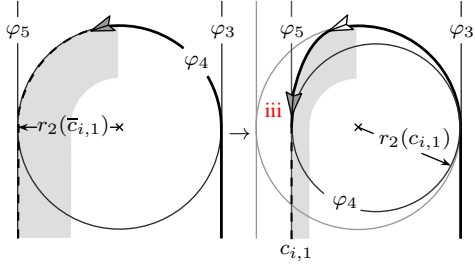

This section solves the problem 
in Sec.~\ref{sec:pbfor}. It provides a plan 
and re-plans-schedules such plan energy-wise. 

\vspace*{-1.4ex}
\subsection{Coverage}
\label{sec:cov-algo}

There are various approaches in the literature to solve CPP problems (e.g., Sec.~\ref{sec:pbfor}). Those that ensure 
completeness 
are NP-hard~\cite{arkin2000approximation} and use cellular decomposition, dividing the free space into sub-regions to be easily covered~\cite{choset2001coverage,galceran2013survey}.

An intuitive way to solve the problem is with a back-and-forth motion, sweeping the space delimited by $v$ we term $\mathcal{Q}^v$. Although abundant in both mobile ground-based~\cite{choset2001coverage
} and aerial~\cite{araujo2013multiple,
cabreira2018energy,difranco2015energy} robotics literature, the motion, called \emph{boustrophedon motion}~\cite{choset2001coverage}, is unsuitable for aerial robots broadly, especially for fixed-wing aerial robots. These robots have reduced maneuverability~\cite{dille2013efficient,mannadiar2010optimal,
xu2014efficient} and are generally unable to fly quick turns~\cite{wang2017curvature}.

To address fixed wings and aerial robots generally, this section details a different motion with a wide turning radius. It is similar to another motion in the literature, the \emph{Zamboni motion}~\cite{araujo2013multiple}, but additionally allows variable CPP 
{\color{black} by dynamically altering the distance between the survey lines with the path parameters}. 
{\color{black} Although cover variability is already considered in the literature~\cite{difranco2015energy}, it is limited to boustrophedon motion for rotary wings.}
The novel motion is termed \emph{Zamboni-like motion} and is composed of four primitive paths
: two lines $\varphi_1,\varphi_2$ and two circles $\varphi_3,\varphi_4$.

We assume the vertices $v_1,v_2,\dots$ are ordered from the top-left-most vertex 
clockwise
, the aerial robot can overfly the edges formed by the vertices, and ${}^{v_x}|_{v_y}$ indicates the edge formed by vertices $v_x,v_y$. 
Algorithm~\ref{alg2} details the procedure to generate the plan $\Gamma$ that covers $\mathcal{Q}^v$ 
{\color{black}at }discretized time step{\color{black}s}, i.e., $\mathcal{T}:=\{t_0,t_0+h,\dots,t_f\}$ for a given step $h\in\mathbb{R}_{>0}$. The algorithm assumes that the line parallel to ${}^{v_1}|_{v_{|v|}}$ is always connected. 
{\color{black}C}omplex covering is possible by, e.g., dividing $\mathcal{Q}^{v}$ into cells 
and 
covering each cell~\cite{choset2001coverage}.

\begin{algorithm}[t]
  \begin{algorithmic}[1]
    \small

    \FORALL{$t\in\mathcal{T}$}
      \color{black}\STATE \textbf{if} $\mathbf{p}
      =\mathbf{p}_{\Gamma_l}${ in Definition~\ref{def:trigs}} \textbf{then return }$\Gamma$\vspace*{.3ex}\color{black}

        

      \color{black}\IF{$\mathbf{p}
      =\mathbf{p}_{\Gamma_i}$}\color{black}
        \STATE $i\gets i+1$\vspace*{.3ex}
        \IF{$i\notin[n]_{>0}$}\label{alg2:cond}
          \STATE $i\gets 1$\vspace*{0ex}
          \STATE $\varphi_{|\Gamma|+1}\gets$ line in Definition~\ref{def:paths} 
          parallel to ${}^{v_1}|_{v_{|v|}}$ that\vspace*{.5ex} \hspace*{1em}intersects $\mathbf{p}_{|\Gamma|}$\vspace*{.3ex}

          \STATE $\mathbf{p}_{|\Gamma|+1}\gets$ other intersection 
          of $\varphi_{|\Gamma|+1}$ and $v$\vspace*{.3ex}

          \STATE $\varphi_{|\Gamma|+2}\gets$ circle 
          whose left most point lays on $\mathbf{p}_{|\Gamma|+1}$\vspace*{.3ex}\label{alg2:circ1}
          
          \STATE $\mathbf{p}_{|\Gamma|+2}\gets$ other inter. 
          of $\varphi_{|\Gamma|+2}$ and $v$\vspace*{.3ex}

          \STATE $\varphi_{|\Gamma|+3}\gets$ line 
          par. to $\varphi_{|\Gamma|+1}$ that inter. $\mathbf{p}_{|\Gamma|+2}$\vspace*{.3ex}

          \STATE $\mathbf{p}_{|\Gamma|+3}\gets$ other inter. 
          of $\varphi_{|\Gamma|+3}$ and $v$\vspace*{.3ex}

          \STATE $\varphi_{|\Gamma|+4}\gets$ circle in Eq.\hspace*{.7ex}(\ref{eq:second-circ-gene}) 
          whose right most point\vspace*{.3ex} \hspace*{1em}lays on $\mathbf{p}_{|\Gamma|+3}$\vspace*{.3ex}\label{alg2:circ2}

          \STATE $\mathbf{p}_{|\Gamma|+4}\gets$ other inter. 
          of $\varphi_{|\Gamma|+4}$ and $v$\vspace*{.3ex}\label{alg2:trig4}

          \vspace*{.8ex}
          \STATE $\Gamma\gets\Gamma\cup\{\Gamma_{|\Gamma|+1},\dots,\Gamma_{|\Gamma|+4}\}${ in Definitions~\ref{def:stage}--\hyperref[def:plan]{4}}\label{alg2:last}

        \ENDIF
      \ENDIF
    \ENDFOR
  \end{algorithmic}
  \caption{Zamboni-like motion for CPP}\label{alg2}
\end{algorithm}

To implement the variable CPP, the radius $r_2$ of the second circle $\varphi_{|\Gamma|+4}$ on Line~\ref{alg2:circ2}
\vspace*{-1ex}
\begin{equation}\label{eq:r2}
  {\small r_2(c_{i,1}):=\sqrt{\smash[b]{r^2+c_{i,1}}},}
  \vspace*{-.4ex}
\end{equation}
is expressed as a function of a path parameter $c_{i,1}\in(\underline{r}^2-r^2,0]$, relative to the last circle in each set of primitive stages. $r\in\mathbb{R}_{>0}$ is a given ideal turning radius along with the minimum radius (see Sec.~\ref{pb:cov-pb}). The center also changes
\begin{equation}\label{eq:second-circ-gene}
  \varphi_{|\Gamma|+4}:=(x-x_{\mathbf{p}_{|\Gamma|+3}}+r_2)^2+(y-y_{\mathbf{p}_{|\Gamma|+3}})^2-r_2^2,
\end{equation}
where $(x_\mathbf{p},y_\mathbf{p})=:\mathbf{p}$ for any point $\mathbf{p}$. Fig.~\ref{fig:tee1} illustrates the concept of $c_{i,1}$ altering the CPP. The radius of the first circle on Line~\ref{alg2:circ1} is then $r_1:=r+x_\mathbf{d}/2$ (i.e., the radiuses of the two circles ensure that the primitive paths are shifted of $\mathbf{d}$).

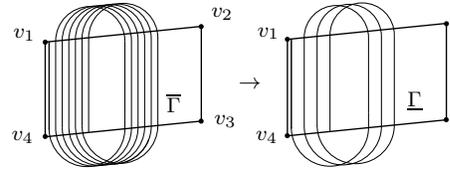
\begin{figure}[t]
  \vspace*{-1ex}
  \footnotesize
  \begin{minipage}[l]{0.7\columnwidth}
    \centering
    \input{\figpath/new/traj3.tikz}
  \end{minipage}\hfill
  \begin{minipage}[l]{0.26\columnwidth}
    \caption{Zamboni-li- ke motion: $\overline{\Gamma}$ with four primitive paths (Lines~\ref{alg2:circ1}--\ref{alg2:trig4} in Algorithm~\ref{alg2}) can be re-planned up to $\underline{\Gamma}$ {\color{black}with the radius} $r_2$.
    }
    \label{fig:zambo}
  \end{minipage}
  \vspace*{-5ex}
\end{figure}

Algorithm~\ref{alg2} initializes $i$ to minus one and builds the first four primitive functions $\varphi_1,\dots,\varphi_4$. The remaining $\Gamma$ is built with the shift $\mathbf{d}$ up to the final point $\mathbf{p}_{\Gamma_l}$. The initial point is $\mathbf{p}_{\Gamma_1}$, placed s.t. the line $\varphi_1$ is at the same distance from an eventual previous line, e.g., $x_{\mathbf{p}_{\Gamma_1}}=x_{v_1}+x_{\mathbf{d}}/2$ in Fig.~\ref{fig:zambo}.

\subsection{Re-planning-scheduling}
\label{sec:repla-algo}

Past literature on planning-scheduling often relies on 
optimization {\color{black} as well as heuristics-based} approaches~\cite{brateman2006energy,zhang2007low,ondruska2015scheduled,lahijanian2018resource}. We similarly derive an optimal control problem {\color{black}and a greedy approach} returning the trajectory of parameters $c_i(\mathcal{T})$ with $\mathcal{T}:=[t_0+h,t_f]$ (see Definition~\ref{def:meas}). Since the final time 
and the 
value of the state $\mathbf{q}$ are not known, we use 
output model predictive control (MPC) that derives the configuration for a finite horizon on an estimated state $\hat{\mathbf{q}}$, i.e., $t_f:=t_0+N$ for a given $N\in\mathbb{R}_{>0}$. 
{\color{black} We utilize MPC to derive the trajectory of the computation parameters and the greedy approach with heuristics remaining coverage time for the path parameters.}

\begin{algorithm}[t]
  \begin{algorithmic}[1]
    \small
    \FORALL{$t\in\mathcal{T}$}
      \makeatletter
      \setcounter{ALC@line}{15}
      \makeatother
      \color{black}\STATE $\mathbf{q}(\mathcal{K}\setminus\{t+N\}),c_i^\sigma(\mathcal{K})\gets${ \vspace*{.3ex}solve NLP }$\argmax_{\mathbf{q}(k),c_i(k)}$ \vspace*{.7ex}\hspace*{1em}${l_f(\mathbf{q}(t+\hspace*{-.4ex}N),t+\hspace*{-.4ex}N)}+\hspace*{-.5ex}{\sum_{k\in\mathcal{K}}{l_d(\mathbf{q}(k),c_i(k),k)}}${\hspace*{.4ex}in\hspace*{.4ex}Eq.\hspace*{.4ex}(\ref{eq:ocp-output-mpc}) \hspace*{1em}on }$\mathcal{K}=\{t,t+h,\dots,t+N\}$\vspace*{.3ex}\label{alg:mpc}
      
      \vspace*{.8ex}
      \color{black}\STATE $k\gets t$\vspace*{.3ex}\label{alg:bat1}
      \WHILE{$b_d(y(k))>0$}\vspace*{.3ex}
        \IF{$k+h\notin\mathcal{K}$}
          \STATE $\mathbf{q}(k+h)\gets${ solve model in Eq.~(\ref{eq:state-perf-q})}\vspace*{.3ex}\label{alg:evol}
        \ENDIF
        \STATE $b_d(y(k+h))\gets${ solve model in Eq.~(\ref{eq:batdyn})}\vspace*{.3ex}
        \STATE $k\gets k+h$\vspace*{.3ex}
      \ENDWHILE
      \STATE $t_b\gets k-t$\vspace*{.3ex}\label{alg:bat2}
      \vspace*{-2.4ex}
      \color{black}\STATE $t_r\gets (\mathrm{diag}(\nu_i^\rho)\vspace*{.3ex}c_i^\rho(t-h)+\tau_i^\rho)[\overbrace{\begin{matrix}1&1&\cdots&1\end{matrix}}^{\rho}]-t$\vspace*{.3ex}\label{alg:traj1}
      \IF{$t_r>t_b$}
        \color{black}\STATE $c_i^{\rho}(t)\gets${ find }$c_i^{\rho}${ with }$t_r\in[0,t_b]${, otherwise take }$\underline{c}_i^\rho$\vspace*{.3ex}\label{alg:traj2}
      \color{black}\ENDIF
      \vspace*{.8ex}
      \color{black}\STATE $\hat{\mathbf{q}}(t+h)\gets${ estimate }$\mathbf{q}${ in Eq.~(\ref{eq:state-perf-q}) with energy sensor }$\Upsilon(t)$\vspace*{-1.8ex}\label{alg:klm1}
      \color{black}\STATE $\hat{y}(t+h)\gets${ derive }$y${ from Eq.~(\ref{eq:state-perf-y}) with est. state }$\hat{\mathbf{q}}(t+h)$
      \label{alg:klm2}
    \ENDFOR
  \end{algorithmic}
  \caption{Coverage re-planning-scheduling}\label{alg}
\end{algorithm}

\vspace*{.4ex}
An optimal control problem (OCP) that selects 
{\color{black} $c_i^\sigma$} 
\begin{subequations}\small\label{eq:ocp-output-mpc}\begin{align}
  \max_{\mathbf{q}(t),c_i(t)}&{l_f(\mathbf{q}(t_f),t_f)+\int_{t_0}^{t_f}{l(\mathbf{q}(t),c_i(t),t)\,dt}},\label{eq:ocp-costs}\\
  \text{s.t. }\dot{\mathbf{q}}&=f(\mathbf{q}(t),c_i(t),t),\label{eq:dyn-evol}\\
  \mathbf{q}(t)&\hspace*{-.2ex}\in\hspace*{-.2ex}\mathbb{R}^m,\,y(t)\hspace*{-.2ex}\in\hspace*{-.2ex}\mathcal{Y}(t)\text{, i.e., constraint in Eq.~(\ref{eq:output-const})},\label{eq:batt-const-mpc}\\
  c_{i,j}(t)&\hspace*{-.2ex}\in\hspace*{-.2ex}\mathcal{C}_{i,j},\,c_{i,\rho+k}(t)\hspace*{-.2ex}\in\hspace*{-.2ex}\mathcal{S}_{i,k}\,\forall j\hspace*{-.2ex}\in\hspace*{-.2ex}[\rho]_{>0},\,k\hspace*{-.2ex}\in\hspace*{-.2ex}[\sigma]_{>0},\label{eq:state-cont-const-mpc}\\
  \mathbf{q}(t_0)&=\hat{\mathbf{q}}_0\,\,\,\text{given (last estimated state)},\text{ and}\label{eq:ocp-outp-mpc-state-est}\\
  b(t_0)&=b_0\,\,\,\text{given}\label{eq:ocp-outp-bat},
\end{align}\end{subequations}
where $\mathbf{q}(t)$ and $c_i(t)$ are the state and parameters trajectories and $l:\mathbb{R}^m\times\mathcal{C}_i\times\mathcal{S}_i\times\mathbb{R}_{\geq 0}\rightarrow\mathbb{R}$ is a given initial cost function with the quadratic expression
\vspace*{-.5ex}
\begin{equation}\label{eq:insta-cost-mpc}
  l(\mathbf{q}(t),c_i(t),t)=\mathbf{q}'(t)Q\mathbf{q}(t)+c_i'(t)Rc_i(t),
  \vspace*{-.5ex}
\end{equation}
where $Q\in\mathbb{R}^{m\times m},R\in\mathbb{R}^{n\times n}$ are given positive semidefinite matrices. 
The final cost function $l_f:\mathbb{R}^m\times\mathbb{R}_{> 0}\rightarrow\mathbb{R}$ is also a quadratic expression 
\vspace*{-.5ex}
\begin{equation}\label{eq:final-cost-mpc}
  l_f(\mathbf{q}(T),T)=\mathbf{q}'(T)Q_f\mathbf{q}(T), 
  \vspace*{-.5ex}
\end{equation}
where $Q_f\in\mathbb{R}^{m\times m}$ is a given positive semidefinite matrix.

Eq.~(\ref{eq:dyn-evol}) is the model in Eq.~(\ref{eq:state-perf}). {\color{black}It} requires a value of the period $T$, which is 
the time needed to fly the four primitive paths in the Zamboni-like motion, i.e., the time 
between two positive evaluations of the condition on Line~\ref{alg2:cond}.

Eq.~(\ref{eq:state-cont-const-mpc}) are the parameters constraints sets in Definition~\ref{def:stage}. Eq.~(\ref{eq:batt-const-mpc}) are the state and output constraints 
that evolve the battery model in Eq.~(\ref{eq:batdyn}). Eq.~(\ref{eq:ocp-outp-mpc-state-est}) is the state guess estimated via state estimation 
(
first estimate is given). Eq.~(\ref{eq:ocp-outp-bat}) is the initial battery SoC from, e.g., flight controller.

Line~\ref{alg:mpc} in Algorithm~\ref{alg} contains a transcribed version of the OCP in Eq.~(\ref{eq:ocp-output-mpc}) into a nonlinear program (NLP) that can be 
solved with available NLP solvers
. Its solution leads to both trajectories of {\color{black} computation} parameters and states for future $N$ instants. Here, the sets $\mathcal{K},\mathcal{T}$ have possibly different steps $h$ (not to be confused with the altitude){\color{black}: 
the set $\mathcal{K}$ is used for the numerical simulation, whereas $\mathcal{T}$ is for re-planning, meaning that $h$ tunes the precision and the frequency of re-planning for $\mathcal{K}$ and $\mathcal{T}$ respectively.}
The functions $l_d,b_d$ are the discretized versions of Eq.~(\ref{eq:insta-cost-mpc})~and~(\ref{eq:batdyn}).

Lines~\ref{alg:bat1}--\ref{alg:bat2} estimate the time needed to completely drain the battery, exploiting the SoC already predicted previously on Line~\ref{alg:mpc}. The {\color{black}path parameters and thus the} coverage is then re{\color{black}-}planned 
on Lines~\ref{alg:traj1}--\ref{alg:traj2} using {\color{black} the heuristics with the} 
scaling factors from Eq.~(\ref{eq:scale-traj}) with $c_{i}^{\rho}(t_0)$ given. 
{\color{black} Concretely, these lines implement the greedy approach by decreasing the path parameters of a given value $\delta_i$ or similarly increasing the parameters when $t_r\leq t_b$ within the bounds (this latter analogous case is not shown explicitly in Algorithm~\ref{alg2} but implemented in Sec.~\ref{sec:experimental})}.
Lines~\ref{alg:klm1}--\ref{alg:klm2} estimate the 
state with 
energy sensor reading $\Upsilon$, {\color{black}using}, e.g., 
Kalman filter
.

Algorithm~\ref{alg} implements Eq.~(\ref{eq:ocp-output-mpc}) for the purpose of energy-aware re-planning-scheduling of $\Gamma$ from Algorithm~\ref{alg2}, i.e, Lines~\ref{alg:mpc}--\ref{alg:klm2} continue after Line~\ref{alg2:last} in Algorithm~\ref{alg2}.

\begin{figure*}
  \centering
  \footnotesize
  \begin{minipage}[t]{0.63\columnwidth}
    \input{\figpath/results/new_physics2/new_trajs-ener_I-II.tikz}
  \end{minipage}\hfill
  \begin{minipage}[t]{0.36\columnwidth}
    \vspace*{-53.8ex}
    \centering
    \caption{\color{black}Results for path, energy, and energy models of two boundary configurations, the lowest \hyperref[fig:trajs-I-static]{I} and the highest \hyperref[fig:trajs-II-static]{II}, without energy-aware planning-scheduling. The use case is that of coverage path planning with Zamboni-like motion and ground hazards detection scheduling. In \hyperref[fig:stat]{a} are the trajectories of the coverage. In \hyperref[fig:stat]{b} are the energy and the period evolutions for both \hyperref[fig:trajs-I-static]{I} and \hyperref[fig:trajs-II-static]{II} with different atmospheric conditions (i.e., different wind speed and direction) and initial guesses, and in \hyperref[fig:stat]{c} the states of the energy model for \hyperref[fig:trajs-I-static]{I}.}
    \label{fig:stat}
  \end{minipage}
  \vspace*{-2ex}
\end{figure*}
\begin{figure*}
  \centering
  \footnotesize
  \begin{minipage}[t]{0.63\columnwidth}
    \input{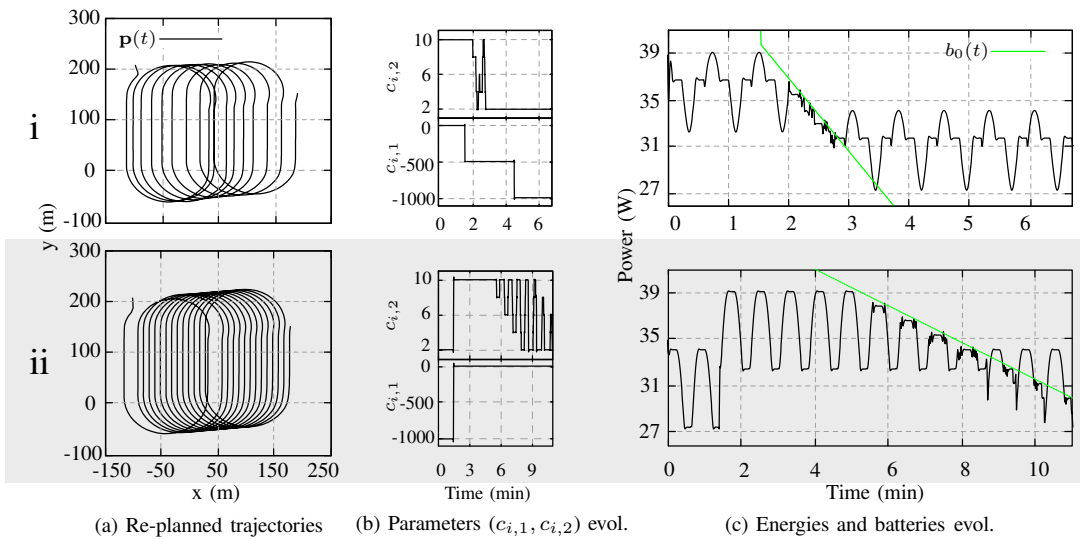}
  \end{minipage}\hfill
  \begin{minipage}[t]{0.36\columnwidth}
    \vspace*{-53.1ex}
    \centering
    \caption{Energy-aware plan- ning-scheduling using the lowest configuration \hyperref[fig:trajs-I-static]{I} as a starting point in \hyperref[fig:trajs-dyn-i]{i} and the highest \hyperref[fig:trajs-II-static]{II} in \hyperref[fig:trajs-dyn-ii]{ii} while varying atmospheric (same as Fig.~\ref{fig:stat}) and battery conditions. In \hyperref[fig:dyn]{a} are the re-planned trajectories, showing the re-planning in the proximity of simulated battery drops. In \hyperref[fig:dyn]{b} are the parameters evolutions, and in \hyperref[fig:dyn]{c} the energy w.r.t. the battery.}
    \label{fig:dyn}
  \end{minipage}
  \vspace*{-3ex}
\end{figure*}

\section{Numerical Simulations} %
\label{sec:experimental}        %
                                %
\vspace*{-.2ex}

Numerical simulations of Algorithms~\ref{alg2}--\ref{alg} in this section are implemented in \textsc{Matlab}\hspace{.5ex}(R) and are extended with the computations energy model on NVIDIA\hspace{.5ex}(R) Jetson Nano\hspace{.5ex}(TM) heterogeneous computing hardware. These simulations complement early data of physical flights of a static coverage plan with the open-source Paparazzi flight controller
.
The computing hardware carries a camera as a peripheral and is evaluated independently of the aerial robot with \powprof{} (see Sec.~\ref{sec:mod-com}). The scheduler, implemented using the Robot Operating System (ROS) middleware
, varies a computation parameter $c_{i,2}$ relative to the ground patterns detection rate from two to ten frames per second (FPS). The detection uses PedNet, a Convolutional Neural Network (CNN)~\cite{ullah2018pednet}, 
implemented in ROS. The planner varies the path parameter $c_{i,1}$ 
between zero and -1000 (i.e., the planner-scheduler is the concrete implementation of Algorithms~\ref{alg2}--\ref{alg}). The set of parameters is unaltered 
through{\color{black}out} the flight, i.e, $c_i:=\begin{bmatrix}c_{i,1}&c_{i,2}\end{bmatrix}',\forall i${\color{black}, along with $\delta_i$ 
in the greedy approach}. 

Fig.~\ref{fig:il-abs} details the data of the physical flight  
in standard atmospheric conditions. 
Fig{\color{black}}.~\ref{fig:stat}{\color{black}--\ref{fig:dyn}} extends the flight with the computing hardware aided by a flight simulation implemented in \textsc{Matlab}\hspace{.5ex}(R). 
Upper-case roman numerals I,II indicate the plans are static (i.e., solely Algorithm~\ref{alg2}), lower-case i,ii exploit planning-scheduling. 

Fig{\color{black}}.~{\color{black}\hyperref[fig:stat]{6a}--\hyperref[fig:dyn]{7a}} 
illustrate the same plan $\Gamma$ under different conditions. Flights \hyperref[fig:trajs-I-static]{I}--\hyperref[fig:trajs-dyn-i]{i} have a constant wind speed of five meters per second, a wind direction of zero degrees, and initial parameters $c_{i,1},c_{i,2}$ values of zero and ten (i.e., full $r_2$ and detection). Flights \hyperref[fig:trajs-II-static]{II}--\hyperref[fig:trajs-dyn-ii]{ii} (see added gray background for clarity) are the same but a wind direction of 90 degrees and the initial parameters values of -1000 and two (i.e., minimum $r_2$ and detection). 
{\color{black} The initial values of path and computation parameters are chosen to represent the highest and lowest configurations in the search space in \hyperref[fig:trajs-I-static]{I}--\hyperref[fig:trajs-dyn-i]{i} and \hyperref[fig:trajs-II-static]{II}--\hyperref[fig:trajs-dyn-ii]{ii} respectively, modeling the behavior of the best- and worst-case scenarios.
Different search strategies are possible by, e.g., running an ideal instance of planning-scheduling prior to the flight.}

Fig{\color{black}}.~
{\color{black}\hyperref[fig:stat]{6b}--\hyperref[fig:dyn]{7c}} 
illustrates first the power ($\Upsilon$ on Line~\ref{alg:klm1} in Algorithm~\ref{alg}), and then the energy model ($y$ on Line~\ref{alg:evol}).
{\color{black} Fig.~\hyperref[fig:stat]{6b}} 
details then the energy model's estimate (
{\color{black}${\hat{y}}$}) on an initial slice
, power ($\Upsilon$), and period ($T$).
{\color{black} Fig.~\hyperref[fig:stat]{6c}} illustrates the evolutions of the state $\mathbf{q}$ in time {\color{black}for \hyperref[fig:ener:static-I]{I}}, concluding that approximately two periods suffice for a consistent 
estimate. 

Flight~\hyperref[fig:ener-dyn-i]{i} simulates a battery ({\color{black}see} green line {\color{black}in Fig.~\hyperref[fig:dyn]{7c}}, the battery behavior $b_0$) drop at approximately one minute and a half and four minutes and a half. Planner-scheduler optimizes the path in the proximity of the drops to ensure that the flight is completed, whereas it maximizes the parameter $c_{i,2}$ {\color{black}(see Fig.~\hyperref[fig:dyn]{7b})} when the battery is discharging, respecting the output constraint
. Flight~\hyperref[fig:ener-dyn-ii]{ii} simulates the opposite scenario: the lowest configuration of parameters and no battery defects. The path parameter increases as soon as the algorithm has estimated enough data (two periods $T$) 
and the computation parameter decreases matching the battery discharge rate. 

{\color{black}
The performance metric is $\Sigma_{t\in\mathcal{T}}(${\small $(w_1\tilde{c}_{i,1}(t)+w_2\tilde{c}_{i,2}(t)$}$)/$ $(${\small $|\mathcal{T}|\text{SoC}(t_f)$}$)$ with {\small $\tilde{c}_{i,j}\hspace*{-.4ex}:=\hspace*{-.4ex}(c_{i,j}\hspace*{-.3ex}-\hspace*{-.3ex}\underline{c}_{i,j})/(\overline{c}_{i,j}\hspace*{-.3ex}-\hspace*{-.3ex}\underline{c}_{i,j})$}. If the initial battery SoC 
is seventy percent and both the parameters are weighted equally, i.e., {\small $w_1\hspace*{-.3ex}=\hspace*{-.3ex}w_2\hspace*{-.3ex}=\hspace*{-.3ex}$} one half, \hyperref[fig:trajs-I-static]{I} would not be able to complete the flight, and \hyperref[fig:trajs-II-static]{II} has a performance metric of zero (i.e., the lowest configuration of parameters throughout the flight).
Nonetheless, performance metrics of \hyperref[fig:trajs-dyn-i]{i} and \hyperref[fig:trajs-dyn-ii]{ii} are 13.05 and 2.24, whereas the average detection and coverage quality is approx. 45 and 35 percent for \hyperref[fig:trajs-dyn-i]{i}, and 62 and 87 percent for \hyperref[fig:trajs-dyn-ii]{ii}.}
For both cases, scaling factors are derived empirically 
{\color{black}similarly to $\delta_i$ set to two hundred fifty}, 
the horizon $N$ is set to six seconds {\color{black} as in} relevant literature~\cite{gavilan2015iterative,
stastny2018nonlinear
}, order $r$ is three
, and the matrices $Q,R,Q_f$ are chosen such that the cost is merely 
squared control. {\color{black} $h$ is set to one-hundredth of a second and to one second for $\mathcal{K}$ and $\mathcal{T}$ respectively to allow sufficient precision and re-planning online.}

{\color{black}
Additional results are reported~\cite{seewaldphdthesis} utilizing simulation capabilities of the Paparazzi flight controller. Data are split into two sets of four flights each, one similar to \hyperref[fig:trajs-dyn-i]{i} and the other to \hyperref[fig:trajs-dyn-ii]{ii}, i.e., initial parameters are at boundary configurations.
These results have an average performance metric of 1.81 and 1.24 for flights similar to \hyperref[fig:trajs-dyn-i]{i} and \hyperref[fig:trajs-dyn-i]{ii} respectively.}

Output MPC on Line~\ref{alg:mpc} relies on a software framework for nonlinear optimization called CasADi~\cite{andersson2012casadi
}, and the popular NLP solver IPOPT~\cite{wachter2006implementation}; both are open-source.

\section{Conclusions and Future Directions}  %
\label{sec:conclusion}                       %
This 
paper provides a planning-scheduling approach for autonomous aerial robots. 
The approach compromises two algorithms: one derives a static coverage plan, 
the other re-plans-schedules the plan on a finite horizon via MPC {\color{black} and a greedy approach}. It evolves the state of the energy model while optimizing battery usage and remedying possible defects. The plan compromise multiple stages, where at each stage the aerial robot flies a path and runs the computations, allowing 
extensibility in terms of constructs and approaches.

{\color{black}To enable physical experiments}, we are currently extending the results to a standard flight controller. 
The study of the implications of planning-scheduling on other energy-critical mobile robots {\color{black}merits additional investigation}. Here, our preliminary study led to possible savings~\cite{seewald2020beyond}, in line with relevant literature~\cite{ondruska2015scheduled,lahijanian2018resource}.
Further directions include {\color{black}the use of a purely optimization-based technique, 
the study of different energy models, and multi-agent planning-scheduling.

{\small\bibliographystyle{IEEEtran}  %
\bibliography{energy-planning}}             %

\end{document}

%% file: figures/new/traj1.tikz
\definecolor{c989898}{RGB}{152,152,152}
\definecolor{cDEDEDE}{RGB}{222,222,222}
\definecolor{c2B2B2B}{RGB}{43,43,43}
\definecolor{cFFFFFF}{RGB}{255,255,255}
\definecolor{c9B9B9B}{RGB}{155,155,155}

\def \globalscale {.77000}
\begin{tikzpicture}[y=0.80pt, x=0.80pt, yscale=-\globalscale, xscale=\globalscale, inner sep=0pt, outer sep=0pt]
\path[draw=c989898,line join=round,line width=0.512pt] (72.6617,91.1923) -- (70.0201,91.1894);

  \path[fill=cDEDEDE,line join=round,even odd rule,line width=0.160pt] (110.1910,90.9491) -- (88.2828,90.9491) .. controls (88.2828,56.6808) and (116.0630,28.9008) .. (150.3310,28.9008) -- (150.3310,50.5905) .. controls (128.1520,50.7291) and (110.2120,68.7416) .. (110.1910,90.9491) -- cycle;

  \path[fill=cDEDEDE,line join=round,even odd rule,line width=0.160pt] (87.7560,160.8600) -- (110.1820,160.8600) -- (110.1820,90.7211) -- (87.7562,90.7211) -- (87.7560,160.8600) -- cycle;

  \path[draw=c989898,line join=round,line width=0.512pt] (98.8887,142.6200) .. controls (82.7540,131.4160) and (72.1903,112.7510) .. (72.1903,91.6196) .. controls (72.1903,57.3513) and (99.9704,29.5712) .. (134.2390,29.5712) .. controls (168.5070,29.5712) and (196.2870,57.3513) .. (196.2870,91.6196) .. controls (196.2870,125.8880) and (168.5070,153.6680) .. (134.2390,153.6680) .. controls (128.3590,153.6680) and (122.6700,152.8500) .. (117.2790,151.3220);

  \path[cm={{1.0,0.0,0.0,1.0,(216.0,65.0)}}] (0.0000,0.0000) node[above right] () {$\mathbf{p}(t_3)$};

  \path[cm={{1.0,0.0,0.0,1.0,(101.0,153.0)}}] (0.0000,0.0000) node[above right] () {$\varphi_0$};

  \path[draw=c2B2B2B,line join=round,line width=0.512pt] (201.2920,55.6358) .. controls (208.3120,65.6905) and (212.4290,77.9218) .. (212.4290,91.1148) .. controls (212.4290,125.3830) and (184.6490,153.1630) .. (150.3810,153.1630) .. controls (116.1130,153.1630) and (88.3328,125.3830) .. (88.3328,91.1148) .. controls (88.3328,56.8465) and (116.1130,29.0665) .. (150.3810,29.0665) .. controls (163.3010,29.0665) and (175.2980,33.0152) .. (185.2330,39.7716);

  \path[draw=c2B2B2B,line join=round,line width=0.512pt] (87.9650,3.0382) -- (87.9650,15.0434);

  \path[draw=c2B2B2B,line join=round,line width=0.512pt] (87.9650,31.3100) -- (87.9652,160.6490);

  \path[draw=c2B2B2B,line join=round,line width=0.512pt] (212.5150,3.0567) -- (212.5150,15.0433);

  \path[draw=c2B2B2B,line join=round,line width=0.512pt] (212.5150,31.2933) -- (212.5150,160.6680);

  \path[draw=c2B2B2B,line join=round,line width=0.512pt] (152.7880,93.4603) -- (148.5070,89.1790);

  \path[draw=c2B2B2B,line join=round,line width=0.512pt] (148.5100,93.4575) -- (152.7910,89.1766);

  \path[fill=black,line join=round,line width=0.256pt] (87.1922,150.2480) -- (87.1922,144.9150) -- (88.4722,144.9150) -- (88.4722,150.2480) -- (87.1922,150.2480) -- cycle(87.1922,139.5820) -- (87.1921,134.2480) -- (88.4721,134.2480) -- (88.4721,139.5820) -- (87.1922,139.5820) -- cycle(87.1921,128.9150) -- (87.1921,123.5820) -- (88.4721,123.5820) -- (88.4721,128.9150) -- (87.1921,128.9150) -- cycle(87.1921,118.2480) -- (87.1921,112.9150) -- (88.4721,112.9150) -- (88.4721,118.2480) -- (87.1921,118.2480) -- cycle(87.1920,107.5820) -- (87.1920,102.2480) -- (88.4720,102.2480) -- (88.4720,107.5820) -- (87.1920,107.5820) -- cycle(87.1920,96.9152) -- (87.1920,91.5818) -- (88.4720,91.5818) -- (88.4720,96.9152) -- (87.1920,96.9152) -- cycle(87.7111,86.2079) -- (87.8096,85.4595) -- (88.5098,81.5392) -- (88.6591,80.9010) -- (89.9133,81.1569) -- (89.7640,81.7950) -- (89.0749,85.6529) -- (88.9836,86.3466) -- (87.7111,86.2079) -- cycle(89.9563,75.6921) -- (91.1514,71.7020) -- (91.5801,70.5702) -- (92.7934,70.9779) -- (92.3648,72.1097) -- (91.1938,76.0189) -- (89.9563,75.6921) -- cycle(93.5349,65.5574) -- (95.8279,60.7421) -- (97.0065,61.2415) -- (94.7135,66.0568) -- (93.5349,65.5574) -- cycle(98.5625,56.0993) -- (99.4757,54.5746) -- (101.6640,51.6890) -- (102.7260,52.4027) -- (100.5380,55.2884) -- (99.6923,56.7010) -- (98.5625,56.0993) -- cycle(105.1130,47.5367) -- (108.8330,43.7147) -- (109.8060,44.5468) -- (106.0860,48.3687) -- (105.1130,47.5367) -- cycle(113.0900,40.3674) -- (115.0590,38.8285) -- (117.5980,37.3783) -- (118.3130,38.4398) -- (115.7740,39.8900) -- (113.9470,41.3182) -- (113.0900,40.3674) -- cycle(122.3150,34.6955) -- (127.2850,32.7611) -- (127.8370,33.9159) -- (122.8670,35.8503) -- (122.3150,34.6955) -- cycle(132.4240,31.0541) -- (137.6210,29.8573) -- (137.9980,31.0806) -- (132.8000,32.2775) -- (132.4240,31.0541) -- cycle(142.9680,29.0255) -- (148.2790,28.5460) -- (148.4800,29.8101) -- (143.1690,30.2896) -- (142.9680,29.0255) -- cycle(153.7150,28.5551) -- (159.0200,29.1031) -- (158.9680,30.3821) -- (153.6630,29.8341) -- (153.7150,28.5551) -- cycle(164.3440,30.0292) -- (169.5280,31.2818) -- (169.3120,32.5435) -- (164.1280,31.2910) -- (164.3440,30.0292) -- cycle(174.6230,33.1078) -- (178.8560,34.7919) -- (179.6090,35.2259) -- (179.0500,36.3776) -- (178.2970,35.9437) -- (174.2350,34.3275) -- (174.6230,33.1078) -- cycle(184.2310,37.8871) -- (185.8180,38.8013) -- (185.1020,39.8619) -- (183.6720,39.0389) -- (184.2310,37.8871) -- cycle(87.1922,160.9150) -- (87.1922,155.5820) -- (88.4722,155.5820) -- (88.4722,160.9150) -- (87.1922,160.9150) -- cycle;

  \path[fill=black,line join=round,line width=0.256pt] (206.9570,64.7718) -- (208.8390,69.7621) -- (207.6720,70.2867) -- (205.7900,65.2963) -- (206.9570,64.7718) -- cycle(201.8560,55.2951) -- (204.4010,59.9820) -- (203.3280,60.6793) -- (200.7830,55.9924) -- (201.8560,55.2951) -- cycle;

  \path[draw=black,line join=round,line width=1.024pt] (209.5140,73.3629) .. controls (212.2000,82.0212) and (212.5620,88.4727) .. (212.5620,88.4727) -- (212.5240,137.0230);

  \path[cm={{1.0,0.0,0.0,1.0,(205.0,27.0)}}] (0.0000,0.0000) node[above right] () {$\varphi_3$};

  \path[cm={{1.0,0.0,0.0,1.0,(188.0,51.0)}}] (0.0000,0.0000) node[above right] () {$\varphi_4$};

  \path[draw=c989898,line join=round,line width=0.512pt] (136.3260,93.3392) -- (132.0430,89.0579);

  \path[draw=c989898,line join=round,line width=0.512pt] (132.0460,93.3358) -- (136.3280,89.0567);

  \path[draw=c989898,line join=round,line width=0.512pt] (72.1527,2.8628) -- (72.1526,15.1033);

  \path[draw=c989898,line join=round,line width=0.512pt] (72.1525,31.3162) -- (72.1518,160.4740);

  \path[draw=black,line join=round,line width=1.024pt] (72.0499,91.7728) .. controls (72.0499,60.9444) and (94.5326,35.3671) .. (123.9960,30.5426);

  \path[draw=black,line join=round,line width=1.024pt] (72.0918,91.4697) -- (72.1456,92.0526);

  \path[draw=black,line join=round,line width=1.024pt] (151.3160,6.6839) .. controls (152.1700,10.5238) and (150.8260,13.1390) .. (149.5430,15.4439) .. controls (145.2990,23.0718) and (137.8250,28.1129) .. (122.4280,30.9623);

  \path[draw=black,line join=round,line width=1.024pt] (72.0839,160.9400) -- (72.0843,91.4542);

  \path[cm={{1.0,0.0,0.0,1.0,(83.0,27.0)}}] (0.0000,0.0000) node[above right] () {$\varphi_5$};

  \path[draw=black,fill=cFFFFFF,line join=round,line width=0.512pt] (110.5350,34.0242) -- (129.7460,35.0223) -- (124.7120,30.4390) -- (126.4240,23.1793) -- (110.5350,34.0242) -- cycle;

  \path[draw=black,line join=round,line width=0.512pt] (110.9590,33.9070) -- (124.5850,30.4380);

  \path[draw=black,line join=round,line width=1.024pt] (212.5330,160.8470) -- (212.5340,111.2790);

  \path[draw=black,fill=c9B9B9B,line join=round,line width=0.512pt] (203.8990,60.2508) -- (204.7900,79.4679) -- (209.7550,74.7215) -- (216.2500,75.0004) -- (203.8990,60.2508) -- cycle;

  \path[draw=black,line join=round,line width=0.512pt] (204.2000,61.0667) -- (209.7480,74.6237);

  \path[draw=black,line join=round,line width=0.512pt] (10.6945,128.4830) -- (10.6945,158.0820);

  \path[draw=black,line join=round,line width=0.512pt] (40.0912,157.8430) -- (10.4916,157.8430);

  \path[cm={{1.0,0.0,0.0,1.0,(0.0,171.0)}}] (0.0000,0.0000) node[above right] () {$\mathcal{O}_W$};

  \path[draw=black,line join=round,line width=0.512pt] (11.6279,157.6980) -- (26.2342,145.9500);

  \path[draw=black,line join=round,line width=0.512pt] (48.7816,127.8150) -- (69.5600,111.1030);

  \path[cm={{1.0,0.0,0.0,1.0,(64.0,27.0)}}] (0.0000,0.0000) node[above right] () {$\varphi_1$};

  \path[cm={{1.0,0.0,0.0,1.0,(26.0,143.5)}}] (0.0000,0.0000) node[above right] () {$\mathbf{p}(t_1)$};

  \path[draw=black,fill=cFFFFFF,line join=round,line width=0.512pt] (71.7014,120.9160) -- (77.6160,102.6100) -- (71.3004,105.3110) -- (65.3168,102.7690) -- (71.7014,120.9160) -- cycle;

  \path[draw=black,line join=round,line width=0.512pt] (71.7059,120.0470) -- (71.2722,105.4050);

\path[draw=c2B2B2B,line join=round,line width=0.512pt] (214.3220,91.2615) -- (212.1830,91.2599);

\path[draw=c2B2B2B,line join=round,line width=0.512pt] (88.4214,91.2331) -- (85.7825,91.2330);

\path[cm={{1.0,0.0,0.0,1.0,(104.0,172.0)}}] (0.0000,0.0000) node[above right] () {$\overline{c}_{4,1}$};

\path[cm={{1.0,0.0,0.0,1.0,(79.0,172.0)}}] (0.0000,0.0000) node[above right] () {$\underline{c}_{4,1}$};

\path[cm={{1.0,0.0,0.0,1.0,(47.0,96.0)}}] (0.0000,0.0000) node[above right] () {$\mathbf{p}_{\Gamma_0}$};

\path[cm={{1.0,0.0,0.0,1.0,(95.0,96.0)}}] (0.0000,0.0000) node[above right] () {$\mathbf{p}_{\Gamma_4}$};

\path[cm={{1.0,0.0,0.0,1.0,(217.0,96.0)}}] (0.0000,0.0000) node[above right] () {$\mathbf{p}_{\Gamma_3}$};

\path[line join=round,line width=0.160pt] (140.1650,61.2828) -- (140.1650,73.7323);

\path[cm={{1.0,0.0,0.0,1.0,(103.0,15.0)}}] (0.0000,0.0000) node[above right] () {$\mathbf{p}(t_0)$};

\path[fill=black,line join=round,line width=0.160pt] (8.2954,131.8670) -- (10.6668,129.7760) -- (12.8405,131.8580) -- (10.5598,125.8400) -- (8.2954,131.8670) -- cycle;

\path[fill=black,line join=round,line width=0.160pt] (37.4883,155.4370) -- (39.5801,157.8080) -- (37.4979,159.9820) -- (43.5161,157.7010) -- (37.4883,155.4370) -- cycle;

\path[fill=black,line join=round,line width=0.160pt] (64.8407,111.5150) -- (67.8900,112.3530) -- (67.3394,115.3120) -- (71.1237,110.1060) -- (64.8407,111.5150) -- cycle;

\end{tikzpicture}

%% file: figures/new/traj2.tikz
\definecolor{cDEDEDE}{RGB}{222,222,222}
\definecolor{c989898}{RGB}{152,152,152}
\definecolor{c2B2B2B}{RGB}{43,43,43}
\definecolor{c4D4D4D}{RGB}{77,77,77}
\definecolor{c9B9B9B}{RGB}{155,155,155}
\definecolor{cFFFFFF}{RGB}{255,255,255}

\def \globalscale {.780000}
\begin{tikzpicture}[y=0.80pt, x=0.80pt, yscale=-\globalscale, xscale=\globalscale, inner sep=0pt, outer sep=0pt]
  \path[fill=cDEDEDE,line join=round,even odd rule,line width=0.160pt] (201.2220,21.9446) .. controls (207.0530,19.9914) and (213.0250,18.6307) .. (219.4860,18.5168) -- (219.4860,50.5755) .. controls (204.1730,51.1381) and (191.8220,63.3026) .. (190.9640,78.5388) -- (190.9770,91.1658) -- (190.9640,91.4517) -- (190.9640,91.7417) .. controls (190.9770,92.4700) and (191.0820,93.2275) .. (191.0620,93.9433) -- (191.0670,93.9503) -- (191.0670,95.7219) -- (190.9640,135.8370) -- (190.9640,143.0540) -- (190.9450,143.0540) -- (190.9450,143.1970) -- (190.9340,144.0070) -- (190.9920,147.6980) -- (179.9870,147.7020) .. controls (180.2660,127.6730) and (179.5510,103.6700) .. (180.4090,85.0349) .. controls (180.8280,75.9339) and (179.7760,64.4703) .. (180.3670,62.4135) .. controls (182.5870,54.6955) and (183.8800,37.4526) .. (200.7770,22.1144) -- (201.2220,21.9446) -- cycle;

  \path[draw=c989898,line join=round,line width=0.512pt] (220.5790,80.1234) ellipse (1.7421cm and 1.7421cm);

  \path[draw=black,line join=round,line width=0.512pt] (222.7490,82.1569) -- (218.4680,77.8757);

  \path[draw=c2B2B2B,line join=round,line width=0.512pt] (283.0110,3.4659) -- (283.0110,13.0980);

  \path[draw=c2B2B2B,line join=round,line width=0.512pt] (283.0110,29.0829) -- (283.0100,147.6390);

  \path[draw=black,line join=round,line width=1.024pt] (212.5660,19.2381) .. controls (214.2910,18.4988) and (220.9040,18.5513) .. (220.9040,18.5513) .. controls (254.9950,18.5513) and (282.6310,46.1876) .. (282.6310,80.2786);

  \path[draw=black,line join=round,line width=0.512pt] (218.4680,82.1570) -- (222.7490,77.8763);

  \path[cm={{1.0,0.0,0.0,1.0,(276.0,26.0)}}] (0.0000,0.0000) node[above right] () {$\varphi_3$};

  \path[draw=black,line join=round,line width=1.024pt] (282.6300,147.6500) -- (282.6300,80.0801);

  \path[draw=c2B2B2B,line join=round,line width=0.512pt] (199.9110,120.3470) .. controls (188.0730,111.0310) and (180.4720,96.5743) .. (180.4720,80.3422) .. controls (180.4720,52.2433) and (203.2500,29.4648) .. (231.3490,29.4648) .. controls (259.4480,29.4648) and (282.2270,52.2433) .. (282.2270,80.3422) .. controls (282.2270,108.4410) and (259.4480,131.2200) .. (231.3490,131.2200) .. controls (228.1390,131.2200) and (224.9980,130.9220) .. (221.9520,130.3540);

  \path[draw=c4D4D4D,line join=round,line width=0.512pt] (180.1260,3.3886) -- (180.1260,13.0871);

  \path[draw=c4D4D4D,line join=round,line width=0.512pt] (180.1260,29.0812) -- (180.1260,147.5620);

  \path[draw=black,line join=round,line width=0.512pt] (275.8090,102.8400) -- (258.8970,95.8481);

  \path[draw=black,line join=round,line width=0.512pt] (229.9920,83.8975) -- (220.7780,80.0882);

  \path[draw=black,line join=round,line width=1.024pt] (181.0490,61.3548) -- (180.1790,80.3748);

  \path[draw=black,line join=round,line width=1.024pt] (212.0450,18.8775) .. controls (189.0000,20.8126) and (182.8960,53.6461) .. (180.3720,62.4206);

  \path[draw=black,fill=c9B9B9B,line join=round,line width=0.512pt] (180.2030,73.0283) -- (186.3000,59.5673) -- (181.2460,61.0826) -- (176.8990,58.6252) -- (180.2030,73.0283) -- cycle;

  \path[fill=black,line join=round,line width=0.256pt] (179.5340,137.2550) -- (179.5340,131.9220) -- (180.8140,131.9220) -- (180.8140,137.2550) -- (179.5340,137.2550) -- cycle(179.5340,126.5890) -- (179.5340,121.2550) -- (180.8140,121.2550) -- (180.8140,126.5890) -- (179.5340,126.5890) -- cycle(179.5340,115.9220) -- (179.5340,110.5890) -- (180.8140,110.5890) -- (180.8140,115.9220) -- (179.5340,115.9220) -- cycle(179.5340,105.2550) -- (179.5340,99.9221) -- (180.8140,99.9221) -- (180.8140,105.2550) -- (179.5340,105.2550) -- cycle(179.5340,94.5887) -- (179.5340,89.2553) -- (180.8140,89.2553) -- (180.8140,94.5887) -- (179.5340,94.5887) -- cycle(179.5340,83.9221) -- (179.5340,80.3520) -- (180.8140,80.3520) -- (180.8140,83.9221) -- (179.5340,83.9221) -- cycle(179.5340,147.9220) -- (179.5340,142.5890) -- (180.8140,142.5890) -- (180.8140,147.9220) -- (179.5340,147.9220) -- cycle;

  \path[cm={{1.0,0.0,0.0,1.0,(173.0,26.0)}}] (0.0000,0.0000) node[above right] () {$\varphi_5$};

  \path[cm={{1.0,0.0,0.0,1.0,(173.0,161.0)}}] (0.0000,0.0000) node[above right] () {$c_{i,1}$};

  \path[cm={{1.0,0.0,0.0,1.0,(233.0,94.0)}}] (0.0000,0.0000) node[above right] () {$r_2(c_{i,1})$};

  \path[draw=black,fill=cFFFFFF,line join=round,line width=0.512pt] (202.9180,21.5792) -- (217.6690,20.7018) -- (213.4350,17.6325) -- (214.1230,11.9445) -- (202.9180,21.5792) -- cycle;

  \path[draw=black,line join=round,line width=0.512pt] (203.2900,21.4420) -- (213.3920,17.6188);

  \path[draw=black,line join=round,line width=0.512pt] (180.2830,72.3687) -- (181.2170,61.1556);

  \path[cm={{1.0,0.0,0.0,1.0,(205.0,130.0)}}] (0.0000,0.0000) node[above right] () {$\varphi_4$};

  \path[draw=c989898,line join=round,line width=0.512pt] (158.8570,3.4053) -- (158.8570,147.5780);

\path[draw=c2B2B2B,line join=round,line width=0.512pt] (137.9100,3.3550) -- (137.9100,12.9842);

\path[draw=c2B2B2B,line join=round,line width=0.512pt] (137.9100,29.2243) -- (137.9100,147.5280);

\path[cm={{1.0,0.0,0.0,1.0,(131.0,26.0)}}] (0.0000,0.0000) node[above right] () {$\varphi_3$};

\path[fill=cDEDEDE,line join=round,line width=0.160pt] (14.4062,77.9517) -- (14.4311,77.9517) .. controls (15.3584,45.0479) and (42.0351,18.5869) .. (75.0243,18.0057) -- (75.0243,50.0644) .. controls (59.7108,50.6270) and (47.3600,62.7914) .. (46.5019,78.0276) -- (46.5148,90.6547) -- (46.5019,90.9406) -- (46.5019,91.2305) .. controls (46.5154,91.9589) and (46.6202,92.7164) .. (46.6001,93.4322) -- (46.6054,93.4392) -- (46.6054,95.2108) -- (46.5019,135.3250) -- (46.5019,142.5430) -- (46.4833,142.5430) -- (46.4829,142.6860) -- (46.4723,143.4960) -- (46.5303,147.1870) -- (14.2304,147.1940) .. controls (14.2304,145.5210) and (14.2092,148.8290) .. (14.1991,145.1270) -- (14.1991,143.5850) -- (14.1991,143.2720) -- (14.1991,142.8530) -- (14.1991,142.5470) -- (14.1991,79.9145) -- (14.4062,77.9517) -- cycle;

\path[draw=c2B2B2B,line join=round,line width=0.512pt] (129.2490,49.1209) .. controls (134.5450,58.2282) and (137.5790,68.8141) .. (137.5790,80.1087) .. controls (137.5790,114.2000) and (109.9430,141.8360) .. (75.8515,141.8360) .. controls (41.7605,141.8360) and (14.1242,114.2000) .. (14.1242,80.1087) .. controls (14.1242,46.0178) and (41.7605,18.3814) .. (75.8515,18.3814) .. controls (90.5804,18.3814) and (104.1040,23.5401) .. (114.7150,32.1492);

\path[draw=black,line join=round,line width=0.512pt] (78.0168,82.1404) -- (73.7410,77.8597);

\path[draw=black,line join=round,line width=0.512pt] (73.7417,82.1430) -- (78.0230,77.8620);

\path[draw=c2B2B2B,line join=round,line width=0.512pt] (14.1483,3.3738) -- (14.1483,13.1642);

\path[draw=c2B2B2B,line join=round,line width=0.512pt] (14.1482,29.1032) -- (14.1477,147.5470);

\path[draw=black,line join=round,line width=0.512pt] (75.9233,80.0349) -- (70.0853,80.0349);

\path[draw=black,line join=round,line width=0.512pt] (25.9299,80.0349) -- (22.7254,80.0349) -- (14.0743,80.0349);

\path[cm={{1.0,0.0,0.0,1.0,(7.0,26.0)}}] (0.0000,0.0000) node[above right] () {$\varphi_5$};

\path[draw=black,line join=round,line width=1.024pt] (67.5256,19.0541) .. controls (69.2504,18.3149) and (75.8630,18.3674) .. (75.8630,18.3674) .. controls (90.5596,18.3674) and (104.0570,23.5035) .. (114.6570,32.0787);

\path[draw=black,line join=round,line width=1.024pt] (129.2120,49.0240) .. controls (134.5380,58.1500) and (137.5900,68.7659) .. (137.5900,80.0947);

\path[cm={{1.0,0.0,0.0,1.0,(115.0,45.0)}}] (0.0000,0.0000) node[above right] () {$\varphi_4$};

\path[cm={{1.0,0.0,0.0,1.0,(26.0,86.0)}}] (0.0000,0.0000) node[above right] () {$r_2(\overline{c}_{i,1})$};

\path[fill=black,line join=round,line width=0.256pt] (13.8207,136.7570) -- (13.8434,131.4240) -- (15.1234,131.4300) -- (15.1007,136.7630) -- (13.8207,136.7570) -- cycle(13.8662,126.0910) -- (13.8890,120.7580) -- (15.1689,120.7630) -- (15.1462,126.0960) -- (13.8662,126.0910) -- cycle(13.9117,115.4240) -- (13.9345,110.0910) -- (15.2145,110.0960) -- (15.1917,115.4300) -- (13.9117,115.4240) -- cycle(13.9572,104.7580) -- (13.9800,99.4244) -- (15.2600,99.4299) -- (15.2372,104.7630) -- (13.9572,104.7580) -- cycle(14.0028,94.0911) -- (14.0255,88.7579) -- (15.3055,88.7632) -- (15.2828,94.0966) -- (14.0028,94.0911) -- cycle(14.0483,83.4246) -- (14.0710,78.0912) -- (15.3510,78.0967) -- (15.3283,83.4300) -- (14.0483,83.4246) -- cycle(14.0938,72.7580) -- (14.1004,71.2104) -- (14.1116,71.0855) -- (14.1481,70.9655) -- (14.2078,70.8553) -- (14.2888,70.7596) -- (14.3849,70.6790) -- (14.4954,70.6197) -- (14.6155,70.5838) -- (14.7404,70.5731) -- (14.1975,70.5744) -- (14.4982,69.0562) -- (14.8987,67.3757) -- (16.1499,67.6456) -- (15.7494,69.3261) -- (15.4578,70.7980) -- (14.7404,71.8531) -- (15.3804,71.2159) -- (15.3738,72.7634) -- (14.0938,72.7580) -- cycle(16.4170,62.2067) -- (17.1785,59.9626) -- (18.2993,57.1827) -- (19.5002,57.6257) -- (18.3794,60.4055) -- (17.6396,62.5857) -- (16.4170,62.2067) -- cycle(20.5366,52.3016) -- (20.9681,51.4011) -- (23.1393,47.5992) -- (24.2741,48.1913) -- (22.1029,51.9931) -- (21.7089,52.8155) -- (20.5366,52.3016) -- cycle(26.1127,43.1159) -- (26.8488,42.0408) -- (29.4547,38.8941) -- (30.4784,39.6625) -- (27.8725,42.8091) -- (27.1989,43.7932) -- (26.1127,43.1159) -- cycle(33.1525,34.9744) -- (35.2416,32.9173) -- (37.1822,31.3901) -- (38.0302,32.3489) -- (36.0897,33.8762) -- (34.0974,35.8378) -- (33.1525,34.9744) -- cycle(41.5103,28.1602) -- (46.0623,25.3810) -- (46.7946,26.4308) -- (42.2426,29.2100) -- (41.5103,28.1602) -- cycle(50.9591,23.0882) -- (53.4451,21.9644) -- (56.0105,21.1719) -- (56.4647,22.3686) -- (53.8993,23.1611) -- (51.5588,24.2191) -- (50.9591,23.0882) -- cycle(61.1063,19.5980) -- (61.2085,19.5662) -- (66.4313,18.6317) -- (66.7337,19.8753) -- (61.5109,20.8100) -- (61.5605,20.7946) -- (61.1063,19.5980) -- cycle(13.7752,147.4240) -- (13.7979,142.0910) -- (15.0779,142.0960) -- (15.0552,147.4290) -- (13.7752,147.4240) -- cycle;

\path[draw=black,fill=c9B9B9B,line join=round,line width=0.512pt] (58.4473,21.6514) -- (73.2218,21.3631) -- (69.1136,18.1274) -- (70.0273,12.4713) -- (58.4473,21.6514) -- cycle;

\path[draw=black,line join=round,line width=0.512pt] (58.7658,21.5380) -- (69.0161,18.1332);

\path[draw=black,line join=round,line width=1.024pt] (137.5900,147.4660) -- (137.5900,79.8963);

\path[fill=black,line join=round,line width=0.160pt] (20.5224,82.3688) -- (18.4306,79.9974) -- (20.5128,77.8237) -- (14.4946,80.1043) -- (20.5224,82.3688) -- cycle;

\path[fill=black,line join=round,line width=0.160pt] (272.1280,98.6429) -- (273.0150,101.6780) -- (270.2060,102.7620) -- (276.6230,103.2530) -- (272.1280,98.6429) -- cycle;

\path[cm={{1.0,0.0,0.0,1.0,(164.0,74.0)}}] (0.0000,0.0000) node[above right] () {\ref{sth:iii}};

\path[cm={{1.0,0.0,0.0,1.0,(142.0,87.0)}}] (0.0000,0.0000) node[above right] () {$\rightarrow$};

\end{tikzpicture}

%% file: figures/new/traj3.tikz
\def \globalscale {.820000}
\begin{tikzpicture}[y=0.80pt, x=0.80pt, yscale=-.92*\globalscale, xscale=.92*\globalscale, inner sep=0pt, outer sep=0pt]
\path[draw=black,line join=round,line width=0.512pt] (20.8642,24.8115) -- (118.7080,14.9592) -- (118.7080,74.3650) -- (20.8642,84.2173) -- (20.8642,24.8115) -- cycle;

\path[draw=black,line join=round,line width=0.384pt] (23.2872,24.4960) -- (23.2868,79.6844);

\path[draw=black,line join=round,line width=0.384pt] (27.4231,24.1801) .. controls (27.4231,12.2018) and (37.1342,2.4907) .. (49.1125,2.4907) .. controls (61.0909,2.4907) and (70.8020,12.2018) .. (70.8020,24.1801);

\path[draw=black,line join=round,line width=0.384pt] (70.8020,79.1472) .. controls (70.8020,92.2682) and (60.1654,102.9050) .. (47.0447,102.9050) .. controls (33.9239,102.9050) and (23.2873,92.2682) .. (23.2873,79.1472);

\path[draw=black,line join=round,line width=0.384pt] (27.4062,24.1425) -- (27.4062,78.7861);

\path[draw=black,line join=round,line width=0.384pt] (70.8019,24.1163) -- (70.8020,70.1775);

\path[draw=black,line join=round,line width=0.256pt] (70.8020,70.1107) -- (70.8020,79.2413);

\path[draw=black,line join=round,line width=0.384pt] (35.6779,23.2120) -- (35.6779,77.9830);

\path[draw=black,line join=round,line width=0.384pt] (39.8135,22.9055) -- (39.8136,77.5574);

\path[draw=black,line join=round,line width=0.384pt] (43.9493,22.5103) -- (43.9494,77.1626);

\path[draw=black,line join=round,line width=0.384pt] (48.0851,22.1465) -- (48.0851,81.3569);

\path[draw=black,line join=round,line width=0.384pt] (74.9207,23.6118) -- (74.9209,78.7368);

\path[draw=black,line join=round,line width=0.384pt] (31.5420,23.7911) .. controls (31.5420,11.8128) and (41.2531,2.1017) .. (53.2314,2.1017) .. controls (65.2098,2.1017) and (74.9209,11.8128) .. (74.9209,23.7911);

\path[draw=black,line join=round,line width=0.384pt] (31.5423,23.7650) -- (31.5420,78.4794);

\path[draw=black,line join=round,line width=0.384pt] (23.2872,24.4960) -- (23.2868,79.6844);

\path[draw=black,line join=round,line width=0.384pt] (79.0565,23.3177) -- (79.0567,78.4427);

\path[draw=black,line join=round,line width=0.384pt] (35.6778,23.4077) .. controls (35.6778,11.4293) and (45.3889,1.7182) .. (57.3672,1.7182) .. controls (69.3455,1.7182) and (79.0566,11.4293) .. (79.0566,23.4077);

\path[draw=black,line join=round,line width=0.384pt] (83.1923,22.8332) -- (83.1924,77.9583);

\path[draw=black,line join=round,line width=0.384pt] (39.8135,22.9506) .. controls (39.8135,10.9723) and (49.5246,1.2612) .. (61.5029,1.2612) .. controls (73.4812,1.2612) and (83.1924,10.9723) .. (83.1924,22.9506);

\path[draw=black,line join=round,line width=0.384pt] (79.0567,78.3999) .. controls (79.0567,91.5208) and (68.4202,102.1570) .. (55.2994,102.1570) .. controls (42.1786,102.1570) and (31.5421,91.5208) .. (31.5421,78.3999);

\path[draw=black,line join=round,line width=0.384pt] (83.1924,77.9265) .. controls (83.1924,91.0475) and (72.5559,101.6840) .. (59.4351,101.6840) .. controls (46.3143,101.6840) and (35.6778,91.0475) .. (35.6778,77.9265);

\path[draw=black,line join=round,line width=0.384pt] (85.7727,85.9816) .. controls (82.3601,94.9191) and (73.7067,101.2660) .. (63.5708,101.2660) .. controls (50.4501,101.2660) and (39.8135,90.6295) .. (39.8135,77.5085);

\path[draw=black,line join=round,line width=0.384pt] (87.3280,22.3917) -- (87.3282,77.5167);

\path[draw=black,line join=round,line width=0.384pt] (91.4639,77.1154) .. controls (91.4639,90.2362) and (80.8274,100.8730) .. (67.7066,100.8730) .. controls (54.5858,100.8730) and (43.9493,90.2362) .. (43.9493,77.1154);

\path[draw=black,line join=round,line width=0.384pt] (91.4639,21.9514) -- (91.4640,77.0764);

\path[draw=black,line join=round,line width=0.512pt] (172.6530,23.1022) -- (272.3070,13.0677) -- (272.3070,73.5725) -- (172.6530,83.6069) -- (172.6530,23.1022) -- cycle;

\path[draw=black,fill=black,line join=round,line width=0.512pt] (172.6450,21.8970) .. controls (173.2860,21.8970) and (173.8060,22.4172) .. (173.8060,23.0587) .. controls (173.8060,23.7003) and (173.2860,24.2205) .. (172.6450,24.2205) .. controls (172.0030,24.2205) and (171.4830,23.7003) .. (171.4830,23.0587) .. controls (171.4830,22.4172) and (172.0030,21.8970) .. (172.6450,21.8970) -- cycle;

\path[cm={{1.0,0.0,0.0,1.0,(154.0,23.0)}}] (0.0000,0.0000) node[above right] () {$v_1$};

\path[cm={{1.0,0.0,0.0,1.0,(153.0,88.0)}}] (0.0000,0.0000) node[above right] () {$v_4$};

\path[cm={{1.0,0.0,0.0,1.0,(278.0,11.0)}}] (0.0000,0.0000) node[above right] () {$v_2$};

\path[cm={{1.0,0.0,0.0,1.0,(280.0,77.0)}}] (0.0000,0.0000) node[above right] () {$v_3$};

\path[draw=black,line join=round,line width=0.384pt] (175.1210,22.7809) -- (175.1200,78.9904);

\path[draw=black,line join=round,line width=0.384pt] (183.1970,22.0352) .. controls (183.1970,10.9023) and (192.2230,1.8766) .. (203.3560,1.8766) .. controls (214.4890,1.8766) and (223.5150,10.9023) .. (223.5150,22.0352);

\path[draw=black,line join=round,line width=0.384pt] (223.5150,78.4434) .. controls (223.5150,91.8070) and (212.6810,102.6400) .. (199.3180,102.6400) .. controls (185.9540,102.6400) and (175.1210,91.8070) .. (175.1210,78.4434);

\path[draw=black,line join=round,line width=0.384pt] (223.5140,21.9733) -- (223.5150,69.3075);

\path[draw=black,line join=round,line width=0.384pt] (223.5150,69.2395) -- (223.5150,78.5389);

\path[draw=black,line join=round,line width=0.384pt] (183.1980,21.4732) -- (183.1980,77.7401);

\path[draw=black,line join=round,line width=0.384pt] (191.2740,21.2319) -- (191.2740,77.1984);

\path[draw=black,line join=round,line width=0.384pt] (175.1210,22.7809) -- (175.1200,78.9904);

\path[draw=black,line join=round,line width=0.384pt] (231.5910,21.0884) -- (231.5910,77.6542);

\path[draw=black,line join=round,line width=0.384pt] (231.5910,77.6230) .. controls (231.5910,90.9867) and (220.7580,101.8200) .. (207.3940,101.8200) .. controls (194.0310,101.8200) and (183.1980,90.9867) .. (183.1980,77.6230);

\path[draw=black,line join=round,line width=0.384pt] (191.2740,21.2576) .. controls (191.2740,10.1247) and (200.3000,1.0990) .. (211.4330,1.0990) .. controls (222.5660,1.0990) and (231.5910,10.1247) .. (231.5910,21.2576);

\path[draw=black,line join=round,line width=0.384pt] (239.6680,76.8666) .. controls (239.6680,90.2302) and (228.8350,101.0630) .. (215.4710,101.0630) .. controls (202.1070,101.0630) and (191.2740,90.2302) .. (191.2740,76.8666);

\path[draw=black,line join=round,line width=0.384pt] (239.6680,20.3473) -- (239.6680,76.9131);

\path[draw=black,line join=round,line width=0.384pt] (199.3510,20.3985) .. controls (199.3510,9.2656) and (208.3760,0.2399) .. (219.5090,0.2399) .. controls (230.6420,0.2399) and (239.6680,9.2656) .. (239.6680,20.3985);

\path[draw=black,line join=round,line width=0.384pt] (199.3510,20.5179) -- (199.3510,80.9456);

\path[draw=black,line join=round,line width=0.384pt] (74.9230,78.7915) .. controls (74.9230,91.9124) and (64.2864,102.5490) .. (51.1657,102.5490) .. controls (38.0449,102.5490) and (27.4083,91.9124) .. (27.4083,78.7915);

\path[draw=black,line join=round,line width=0.384pt] (87.3488,77.5084) .. controls (87.3488,90.6294) and (76.7122,101.2660) .. (63.5914,101.2660);

\path[draw=black,line join=round,line width=0.384pt] (43.9493,22.5457) .. controls (43.9493,10.5673) and (53.6604,0.8563) .. (65.6387,0.8563) .. controls (77.6171,0.8563) and (87.3282,10.5673) .. (87.3282,22.5457);

\path[draw=black,line join=round,line width=0.384pt] (48.0851,22.1941) .. controls (48.0851,10.2158) and (57.7962,0.5047) .. (69.7745,0.5047) .. controls (81.7528,0.5047) and (91.4639,10.2158) .. (91.4639,22.1941);

\path[draw=black,fill=black,line join=round,line width=0.512pt] (172.5900,82.3602) .. controls (173.2320,82.3602) and (173.7520,82.8804) .. (173.7520,83.5220) .. controls (173.7520,84.1636) and (173.2320,84.6837) .. (172.5900,84.6837) .. controls (171.9480,84.6837) and (171.4280,84.1636) .. (171.4280,83.5220) .. controls (171.4280,82.8804) and (171.9480,82.3602) .. (172.5900,82.3602) -- cycle;

\path[draw=black,fill=black,line join=round,line width=0.512pt] (272.1850,72.3452) .. controls (272.8270,72.3452) and (273.3470,72.8654) .. (273.3470,73.5070) .. controls (273.3470,74.1486) and (272.8270,74.6687) .. (272.1850,74.6687) .. controls (271.5430,74.6687) and (271.0230,74.1486) .. (271.0230,73.5070) .. controls (271.0230,72.8654) and (271.5430,72.3452) .. (272.1850,72.3452) -- cycle;

\path[draw=black,fill=black,line join=round,line width=0.512pt] (272.3250,12.1252) .. controls (272.9670,12.1252) and (273.4870,12.6453) .. (273.4870,13.2869) .. controls (273.4870,13.9285) and (272.9670,14.4487) .. (272.3250,14.4487) .. controls (271.6830,14.4487) and (271.1630,13.9285) .. (271.1630,13.2869) .. controls (271.1630,12.6453) and (271.6830,12.1252) .. (272.3250,12.1252) -- cycle;

\path[draw=black,fill=black,line join=round,line width=0.512pt] (20.8596,23.6827) .. controls (21.5013,23.6827) and (22.0214,24.2028) .. (22.0214,24.8444) .. controls (22.0214,25.4860) and (21.5013,26.0061) .. (20.8596,26.0061) .. controls (20.2181,26.0061) and (19.6980,25.4860) .. (19.6980,24.8444) .. controls (19.6980,24.2028) and (20.2181,23.6827) .. (20.8596,23.6827) -- cycle;

\path[draw=black,fill=black,line join=round,line width=0.512pt] (20.8363,82.8461) .. controls (21.4779,82.8461) and (21.9980,83.3662) .. (21.9980,84.0078) .. controls (21.9980,84.6494) and (21.4779,85.1696) .. (20.8363,85.1696) .. controls (20.1947,85.1696) and (19.6746,84.6494) .. (19.6746,84.0078) .. controls (19.6746,83.3662) and (20.1947,82.8461) .. (20.8363,82.8461) -- cycle;

\path[draw=black,fill=black,line join=round,line width=0.512pt] (118.6260,13.9161) .. controls (119.2680,13.9161) and (119.7880,14.4362) .. (119.7880,15.0778) .. controls (119.7880,15.7194) and (119.2680,16.2396) .. (118.6260,16.2396) .. controls (117.9850,16.2396) and (117.4640,15.7194) .. (117.4640,15.0778) .. controls (117.4640,14.4362) and (117.9850,13.9161) .. (118.6260,13.9161) -- cycle;

\path[draw=black,fill=black,line join=round,line width=0.512pt] (118.5960,73.2495) .. controls (119.2380,73.2495) and (119.7580,73.7697) .. (119.7580,74.4111) .. controls (119.7580,75.0527) and (119.2380,75.5729) .. (118.5960,75.5729) .. controls (117.9550,75.5729) and (117.4350,75.0527) .. (117.4350,74.4111) .. controls (117.4350,73.7697) and (117.9550,73.2495) .. (118.5960,73.2495) -- cycle;

\path[cm={{1.0,0.0,0.0,1.0,(1.0,24.0)}}] (0.0000,0.0000) node[above right] () {$v_1$};

\path[cm={{1.0,0.0,0.0,1.0,(0.0,88.0)}}] (0.0000,0.0000) node[above right] () {$v_4$};

\path[cm={{1.0,0.0,0.0,1.0,(125.0,11.0)}}] (0.0000,0.0000) node[above right] () {$v_2$};

\path[cm={{1.0,0.0,0.0,1.0,(127.0,78.0)}}] (0.0000,0.0000) node[above right] () {$v_3$};

\path[cm={{1.0,0.0,0.0,1.0,(142.0,54.0)}}] (0.0000,0.0000) node[above right] () {$\rightarrow$};

\path[cm={{1.0,0.0,0.0,1.0,(97.0,69.0)}}] (0.0000,0.0000) node[above right] () {$\overline{\Gamma}$};

\path[cm={{1.0,0.0,0.0,1.0,(248.0,68.0)}}] (0.0000,0.0000) node[above right] () {$\underline{\Gamma}$};

\end{tikzpicture}